\documentclass[sn-mathphys-num]{sn-jnl}

\usepackage{graphicx}%
\usepackage{multirow}%
\usepackage{amsmath,amssymb,amsfonts}%
\usepackage{amsthm}%
\usepackage{mathrsfs}%
\usepackage[title]{appendix}%
\usepackage{xcolor}%
\usepackage{textcomp}%
\usepackage{manyfoot}%
\usepackage{booktabs}%
\usepackage{algorithm}%
\usepackage{algorithmicx}%
\usepackage{algpseudocode}%
\usepackage{listings}%
\usepackage{subcaption}%

\theoremstyle{thmstyleone}%
\theoremstyle{thmstyletwo}%

\theoremstyle{thmstylethree}%

\begin{document}

\title[Article Title]{Improve Contrastive Clustering Performance by Multiple Fusing-Augmenting ViT Blocks}


\author[1]{\fnm{Cheng} \sur{Wang}}\email{23071213303@stu.xidian.edu.cn}

\author*[1]{\fnm{Shuisheng} \sur{Zhou}}\email{sszhou@mail.xidian.edu.cn}

\author[1]{\fnm{Fengjiao} \sur{Peng}}\email{22071213330@stu.xidian.edu.cn}

\author[1]{\fnm{Jin} \sur{Sheng}}\email{jinxichen\_xidian@foxmail.com}

\author[1]{\fnm{Feng} \sur{Ye}}\email{fye@xidian.edu.cn}

\author[2]{\fnm{Yinli} \sur{Dong}}\email{124897080@qq.com}

\affil[1]{\orgdiv{School of Mathematics and Statistics}, \orgname{Xidian University}, \orgaddress{ \city{Xi’an}, \postcode{710126}, \state{Shaanxi}, \country{China}}}

\affil[2]{\orgname{General Education College of Xi'an Eurasia University}, \orgaddress{\city{Xi’an}, \postcode{710065}, \state{Shaanxi}, \country{China}}}


\abstract{In the field of image clustering, the widely used contrastive learning networks improve clustering performance by maximizing the similarity between positive pairs and the dissimilarity of negative pairs of the inputs. Extant contrastive learning networks, whose two encoders often implicitly interact with each other by parameter sharing or momentum updating, may not fully exploit the complementarity and similarity of the positive pairs to extract clustering features from input data. To explicitly fuse the learned features of positive pairs, we design a novel multiple fusing-augmenting ViT blocks (MFAVBs) based on the excellent feature learning ability of Vision Transformers (ViT). Firstly, two preprocessed augmentions as positive pairs are separately fed into two shared-weight ViTs, then their output features are fused to input into a larger ViT. Secondly, the learned features are split into a pair of new augmented positive samples and passed to the next FAVBs, enabling multiple fusion and augmention through MFAVBs operations. Finally, the learned features are projected into both instance-level and clustering-level spaces to calculate the cross-entropy loss, followed by parameter updates by backpropagation to finalize the training process. To further enhance the model’s ability to distinguish between similar images, our input data for the network we propose is preprocessed augmentions with features extracted from the CLIP pretrained model. Our experiments on seven public datasets demonstrate that MFAVBs serving as the backbone for contrastive clustering outperforms the state-of-the-art techniques in terms of clustering performance.}




\maketitle

\section{Introduction}

The contrastive learning is a type of self-supervised representation learning that is based on recognizing the similarities and differences between objects \cite{ref34}. In contrast to traditional k-means\cite{ref29} and its variants \cite{ref53,ref50,ref51,ref52}, which rely on data distance or similarity for direct grouping, these learning networks \cite{ref2,ref3,ref4,ref46,ref65} are typically based on a Siamese-like network \cite{ref1} that encodes two pair of samples through two encoders with the goal of pulling positive pairs (similar pairs) closer and pushing negative pairs (dissimilar pairs) apart. The parameters of the two encoders are typically updated using either the shared parameter technique \cite{ref8,ref33,ref9,ref10,ref11} or the momentum updating technique \cite{ref12,ref13,ref32} to better capture similarities between image features. Among them, the memory bank \cite{ref8,ref33} stores samples to utilize for augmenting the quantity of negative pairs in each batch; the In-batch Negative \cite{ref9,ref10,ref11} regards the remaining samples in the batch as negative pairs, and directly optimizes parameters based on the intra-batch loss in order to improve the consistency of feature expression and computational efficiency. Alternatively, the momentum updating technique \cite{ref12,ref13,ref32} leverages a momentum-updated key encoder to generate the large-scale negative representations for the memory bank to ensure more consistent representations for negative pairs. Jin et al. \cite{ref49} proposes a multi-contrast clustering method that integrates multi-resolution augmentation and a momentum-output queue, leading to enhanced clustering performance. These techniques augment negative pairs, enabling the network to bring similar pairs closer and push dissimilar pairs further apart. 

The features of the two augmentions generated by the same encoder may exhibit significant divergence.  Yet, the aforementioned approaches implicitly establish a connection between these augmentions through the shared parameter technique or the momentum updating technique.  Consequently, these approaches may not fully exploit the complementarity and similarity between the features of the two augmentions. Recently, several explicit fusion techniques have been proposed, such as concatenating the features of the encoder and decoder in image segmentation \cite{ref14}, location and channel attention features element-wise addition \cite{ref18}, and performing multiple fusions at intermediate layers \cite{ref19}. However, all of these methods are based on supervised learning and cannot be directly applied to contrastive clustering networks.

Furthermore, contrastive clustering networks usually learn similarity metrics for single-modal data in latent space, which weakens their ability to distinguish between highly similar or even overlapping datasets. Several multimodal models\cite{ref21,ref48} map multiple modalities to a unified representation space using a contrastive learning framework. These models capture a larger amount and more diverse types of information, and their pre-trained models demonstrate excellent zero-shot performance, enabling them to generalize well to unseen data. Multimodal Feature Fusion for Zero-Shot Learning\cite{ref43} literature also shows that fusing features from different modalities results in better learned representations.

Building on previous studies, we design a noval MFAVBs to explicitly fuse features. In particular,  the two preprocessed augmentions, concatenated with features extracted from the CLIP pre-trained model \cite{ref21}, are separately fed into two ViTs with shared weights. The learned features are then concatenated along the token dimension to be processed by a larger ViT for features augmented. Subsequently, these features are split and fed into the next FAVBs. The FAVBs is iterated multiple times to derive the final extracted features, which are then individually input into an instance projector and a cluster projector for contrastive clustering. The complete contrastive clustering algorithm is referred to as MFAVBs-CC. In MFAVBs-CC, the parameters of two encoders are updated using the simply in-batch Negative technique to ensure consistency of feature extraction, and the entire network is trained in an end-to-end manner to obtain the final clustering result. 

The main contributions of this study are as follows:

$\bullet$  The fusing-augmenting strategy enables the model to simultaneously learn clustering and instance-level information, utilizing the ViT architecture to extract shared features between the two, thereby promoting mutual reinforcement and capturing the commonalities among positive samples. On the other hand, the splitting strategy ensures the independence of the two branches.

$\bullet$  By repeatedly applying the FAVBs module, the model can fully leverage the potential of the features of the intermediate layer, improving its representation capability and improving clustering accuracy.

$\bullet$ ViT can be substituted with alternative encoder models that possess the same input and output feature dimensions in order to accommodate various downstream tasks.

$\bullet$ Combining input data with the features extracted from the CLIP pretrained model can increase the model's ability to discriminate images that exhibit similarities across different classes.

The rest of the paper is structured as follows: Section II deals with the architecture
and related research on contrastive clustering networks and CLIP pretrained models, Section III describes the concrete procedures of MFAVBs-CC, including: network architecture, Multiple Fusing-Augmenting ViT Blocks, contrastive clustering, fusing CLIP features and comparison. In addition, section IV verify the validity of MFAVBs-CC by experiments. Finally, Section VI concludes and prospect.

\section{Related Works}

\subsection{Contrastive clustering networks}

The fundamental principle of contrastive learning \cite{ref10} is to train a model by comparing different instances. The goal is to distinguish between similar (positive) and dissimilar (negative) instances. For example, in image processing, different augmentations of a single image act as 'anchor' and 'positive sample,' while other images in the batch form 'negative samples.' These comparisons help the model learn to extract meaningful features.After selecting appropriate instances, the learning model maps these instances into feature space using an encoder and a projection head. The encoder transforms the input data into high-dimensional vector representations, which can be trained end-to-end using backpropagation. Ultimately, the learned features can be applied to various downstream tasks.

Within contrastive clustering techniques, increasing the number of negative samples is a widely used strategy to learn a greater variety of dissimilar instances. For example,  A Simple Framework for Contrastive Learning of Visual Representations (SimCLR) \cite{ref9} learns feature representations by maximizing consistency between the augmented views of different data samples.  Contrastive Clustering (CC) \cite{ref10} improves on SimCLR by performing contrastive learning at the cluster level and minimizing loss at both instance and cluster levels to achieve superior clustering outcomes. Vision transformer for contrastive clustering (VTCC) \cite{ref11}, which builds upon CC, substitutes the backbone with ViT blocks\cite{ref20} to boost the model's ability to learn feature representations. Mutual-Taught Deep Clustering (MTDC) \cite{ref39} proposes , which unifies unsupervised representation learning and unsupervised classification into a framework while realizing mutual promotion using a novel mutual-taught mechanism. 

However, existing networks devote most of their effort to boosting the feature-extraction encoder while largely overlooking cross-channel feature interaction. This oversight hampers their ability to learn discriminative representations—especially on datasets with high inter-class similarity.In contrast, our approach explicitly fuses the dual branches, fully leveraging their complementary information. This targeted fusion strengthens cross-channel interactions and, in turn, significantly enhances the model’s representational power for contrastive clustering tasks.

\subsection{CLIP pretrained model}

Two important modalities are text and visual information, with some approaches combining image features with traditional text representations to enrich the semantic content of images. For instance, CLIP \cite{ref21} mimics a contrastive learning framework by using a transformer encoder and a ViT encoder to process a large collection of matched image-text pairs, converting them into feature vectors. The goal is to make the features of paired data as similar as possible while making those of unpaired data as different as possible by calculating the cosine similarity of these feature vectors. 

Recent advances in vision–language modelling highlight the effectiveness of injecting semantic priors through dedicated tokens. ClsCLIP\cite{ref58}, showing that replacing a ViT’s initial zero-vector CLS token with the text-side CLS embedding from CLIP guides self-attention toward category-relevant regions and delivers state-of-the-art zero-shot semantic-segmentation accuracy. Visual Prompt Tuning \cite{ref59} takes a parameter-efficient route: a handful of learnable prompt tokens are prepended to the patch sequence while the backbone remains frozen, adding <1 \% extra parameters yet consistently outperforming full fine-tuning across diverse downstream tasks. Complementing these ideas, CoOp/CoCoOp \cite{ref60} convert the textual context words in CLIP prompts into learnable vectors and further condition them on individual images, markedly improving generalisation to unseen classes and cross-domain settings. Collectively, these studies demonstrate that introduction of CLS- or prompt-style tokens carrying rich semantic cues incurs negligible computational cost, accelerates convergence, and enhances representational discriminativeness—providing solid theoretical and empirical support for our use of the CLIP CLS embedding as a “pseudo-CLS” initializer in Section \ref{FMF}.

\section{The Proposed Method}

\begin{figure}[h]
  \includegraphics[width=1.0\textwidth]{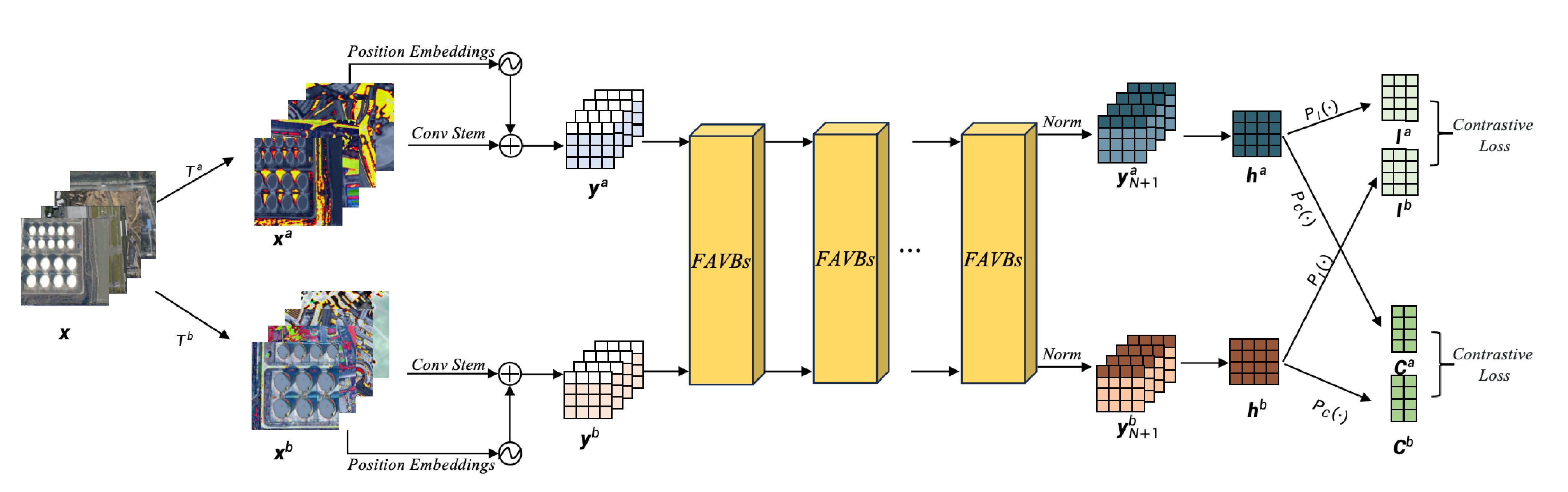}
  \caption{\small The framework of Contrastive Clustering with Multiple Fusing-Augmenting ViT Blocks (MFAVBs-CC). The flowchart includes four pairs randomly selected remote sensing augmentions from the RSOD dataset.   In the diagram, the matrix's square column vector represents the token dimension, and the row vector represents the embedding.}
  \label{fig1}
\end{figure}

The majority of contrastive learning frameworks \cite{ref9,ref10,ref11} are based on a dual-tower architecture. To fully mine the complementarity between different augmented samples to further improve the clustering performance, we propose a explicit feature fusion method within contrastive clustering networks. The complete framework of MFAVBs-CC is illustrated in Figure \ref{fig1}, encompassing Multimodal Features Fusing, MFAVBs, and Contrastive Clustering.

\subsection{Network Architecture}

First, each augmentation echnique is independently applied to a batch of images $\boldsymbol{x}=(x_1,...,x_B)$ (In this paper, the bolded letters represent a set of elements from a batch, while the non-bolded letters denote a specific element of that set. Further explanations will not be provided.) with a specified probability, in accordance with the settings of SimCLR \cite{ref9}. Each image results in two different augmentions through two different combinations of augmentions echniques: $\boldsymbol{x}^a, \boldsymbol{x}^b = \mathcal{T}^a(\boldsymbol{x}), \mathcal{T}^b(\boldsymbol{x})$, where $\mathcal{T}^a$ contains ResizedCrop, ColorJitter, Grayscale, HorizontalFlip, GaussianBlur, $\mathcal{T}^b$ contains ResizedCrop, ColorJitter, Grayscale, HorizontalFlip, GaussianBlur, Solarize. Next, $x^a$ and $x^b$ are passed through a Conv Stem network \cite{ref11} to convert them into 1D image sequences, which are combined with position embeddings to form the original input data ${\boldsymbol{y}^a}$ and ${\boldsymbol{y}^b} \in \mathbb{R}^{B \times (P + 1) \times E}$, where $B$ is the batch size, $P$ is the number of patches squared, and $E$ is the number of embedding dimension. The three aforementioned processes (image Augmentation, Conv Stem, and Positional Embeddings) are collectively referred to as the data preprocessing stage, denoted as $D_a(\cdot), D_b(\cdot)$. These samples ${\boldsymbol{y}^a}$ and ${\boldsymbol{y}^b} \in \mathbb{R}^{B \times (P + 1) \times E}$ are then fed into the improved MFAVBs module to learn feature representations. After normalization, two sets of features $\textbf{$\boldsymbol{h}^a$}$,$\textbf{$\boldsymbol{h}^b$}\in \mathbb{R}^{B \times 1 \times E}$ are obtained and separately input into the instance projector and cluster projector to perform instance-level contrastive learning and global cluster structure learning. The entire framework is trained in an end-to-end approach using two contrastive loss functions to obtain clustering results for downstream tasks. The detailed pseudocode of MFAVBs-CC is presented in Algorithm \ref{Al1}.

\begin{algorithm}[t]
    \caption{MFAVBs-CC:  A PyTorch Pseudo Code}
    \label{Al1}
    \begin{algorithmic}
    \renewcommand{\thealgorithm}{}
    \State $ D_a(\cdot), D_b(\cdot) $: Data Preprocessing
    \State $f_i(\cdot)$: The i-th layer of ViT block
    \State $Cat(\cdot, \cdot)$: Concatenate along the patch dimension
    \State $Spl(\cdot)$: Split along the patch dimension
    \State $ P_I, P_C $: Instance and Cluster Projectors
    \State $ \mathcal{L}_I(\cdot,\cdot), \mathcal{L}_C(\cdot,\cdot)$ : Instance-level and Cluster-level contrastive loss
        \For{$\boldsymbol{x}$ in loader}  \textit{\# $\boldsymbol{x}$ is a batch of images}
            \State $\boldsymbol{y}^a, \boldsymbol{y}^b = D_a(\boldsymbol{x}), D_b(\boldsymbol{x}) $
            \State $ \boldsymbol{y}_1^a, \boldsymbol{y}_1^b = \boldsymbol{y}^a, \boldsymbol{y}^b $
            \For{\( i = 1 \) to \( N \)}  \textit{\#FAVBs stacks N times}
                \State \( \boldsymbol{y}_i^a, \boldsymbol{y}_i^b = f_{2i-1}(\boldsymbol{y}_{i}^a),  f_{2i-1}(\boldsymbol{y}_{i}^b)\)
                \State \( \boldsymbol{y}_i^{new} = Cat(\boldsymbol{y}_i^a,  \boldsymbol{y}_i^b)\)
                \State \( \boldsymbol{y}_i^{newout} = f_{2i}(\boldsymbol{y}_i^{new})\)
                \State \( \boldsymbol{y}_{i+1}^a, \boldsymbol{y}_{i+1}^b = Spl(\boldsymbol{y}_i^{newout})\)
            \EndFor  
            \State $ \boldsymbol{h}^{a}, \boldsymbol{h}^{b} = \boldsymbol{y}^a_{N+1}[: , 0, :], \boldsymbol{y}^b_{N+1}[: , 0,:] $ \textit{\# Extract features}
            \State $ \boldsymbol{I}^a, \boldsymbol{I}^b = P_I(\boldsymbol{h}^a), P_I(\boldsymbol{h}^b) $ \textit{\#Project to instance subspace}
        
            \State $ \boldsymbol{C}^a, \boldsymbol{C}^b = P_C(\boldsymbol{h}^a), P_C(\boldsymbol{h}^b) $  \textit{\#Project to cluster subspace}
            \State $ \mathcal{L} = \mathcal{L}_I(\boldsymbol{I}^a,\boldsymbol{I}^b) + \mathcal{L}_C(\boldsymbol{C}^a,\boldsymbol{C}^b) $ \textit{\# Loss calculation}
            \State loss.backward()
            \State update(${f_1:f_8}$) \textit{\# Optimize the parameters of all ViT blocks} 
        \EndFor
    \end{algorithmic}
\end{algorithm}

\subsection{Multiple Fusing-Augmenting ViT Blocks}\label{MFAVBs}

According to \cite{ref54}, previous methods that rely on simple fusion at the final output layer are unable to effectively align and integrate the information from cluster-level and instance-level, which limits the potential performance gains. The multi-head attention mechanism in ViT captures the internal dependencies among data vectors, allowing for a more efficient integration of the relationships between instance and cluster. The multi-layer design enhances the model’s capacity for abstraction, enabling it to extract deeper image features, thus better stimulating the contrastive learning framework and assisting the model in learning the common features across various augmented images. 

\begin{figure}[htbp]
    \centering
    
    \begin{subfigure}[t]{0.3\textwidth} 
        \centering
        \includegraphics[width=\linewidth]{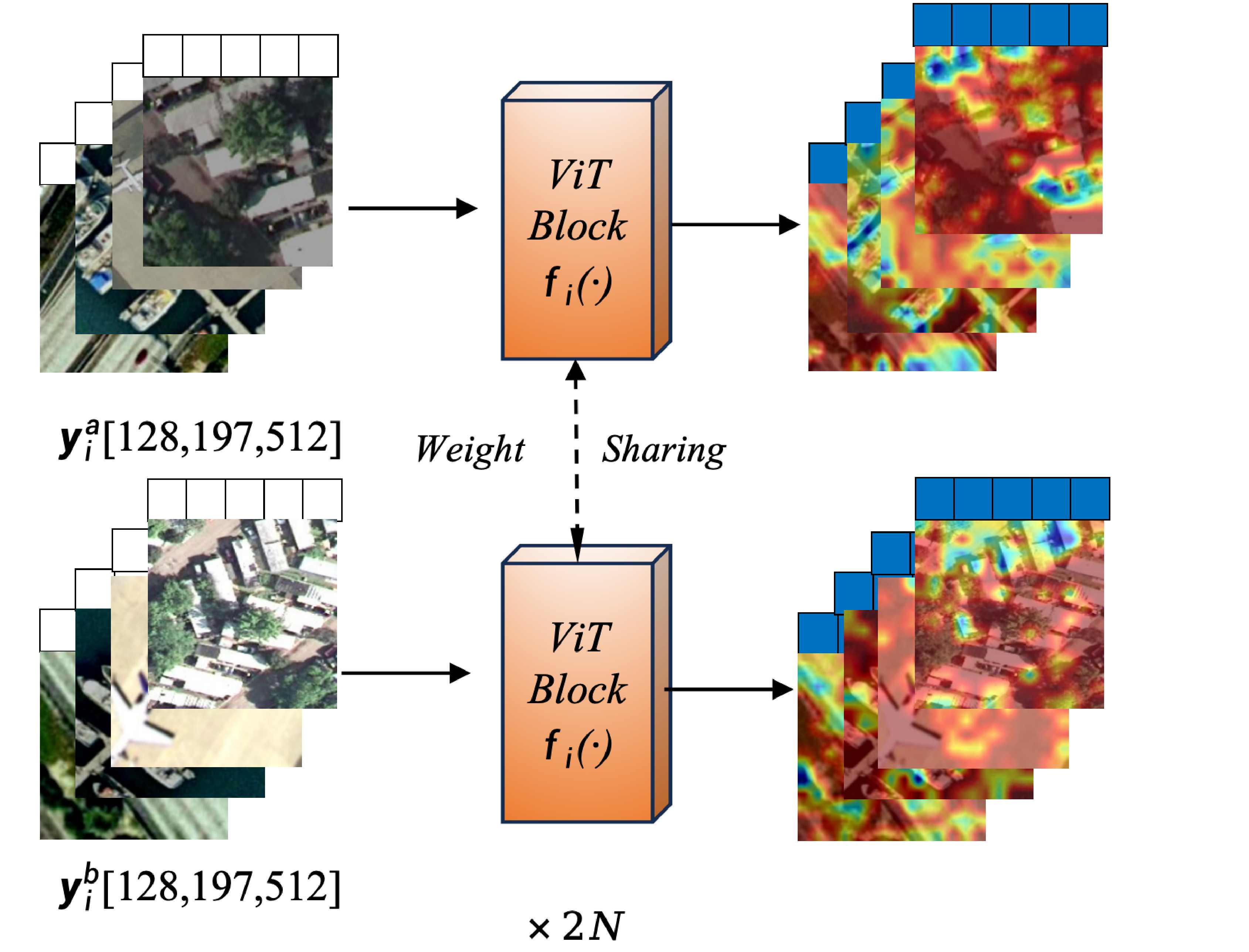}
        \caption{VTCC}
    \end{subfigure}
    \hfill 
    \begin{subfigure}[t]{0.6\textwidth} 
        \centering
        \includegraphics[width=\linewidth]{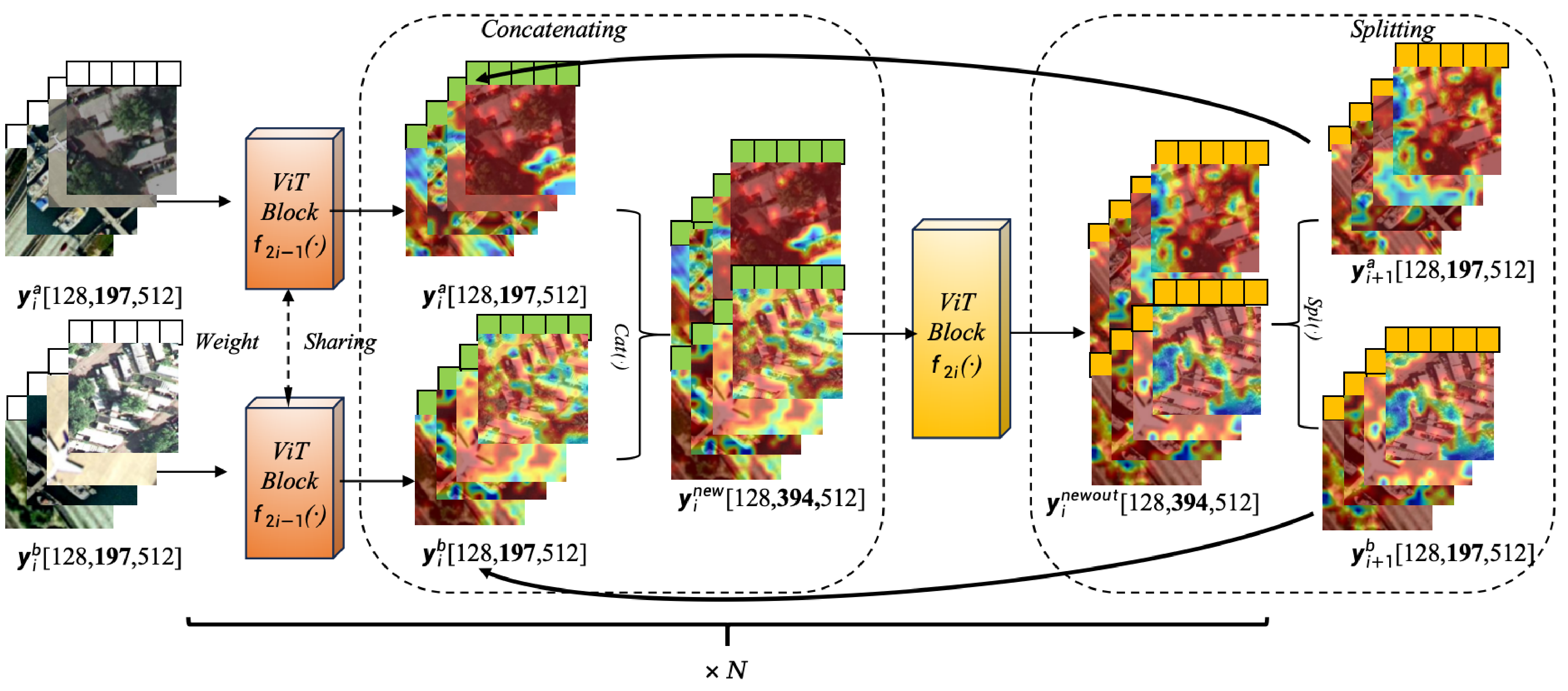}
        \caption{MFAVBs}
    \end{subfigure}
    \caption{\small (a) The original VTCC intermediate layer framework; (b) The intermediate layer framework of the MFAVBs we propose.}
    \label{fig2}
\end{figure}

To better distinguish the source of the fused features and facilitate their separation, we concatenate the features of two augmentions along the token dimension by taking full advantage of the ability of ViT to capture global information associations to better exploit the complementary and similarity relationship between them. Although each ViT block shares the same framework, the parameters differ. Therefore, we denote the ViT block at the i-th layer as $f_i(\cdot)$ to distinguish them. Specifically, the input data $\boldsymbol{y}^a$ and $\boldsymbol{y}^b$ are fed into two ViT blocks with shared parameters for learning, resulting in the learned features $\boldsymbol{y}^a_1, \boldsymbol{y}^b_1 = f_1(\boldsymbol{y}^a), f_1(\boldsymbol{y}^b)$. These features are then concatenated along the token dimension to form the input $\boldsymbol{y}^{new}_1 \in \mathbb{R}^{B \times 2T \times E}$ for the subsequent layer. In this way, while preserving the information of the original input pairs, the fusion of  complementarity and similarity in the features of two augmentions can be effectively achieved by leveraging  characteristics of ViT.  Subsequently, the corresponding branch features can be easily split during the splitting manipulate. The specific concatenating ($Cat(\boldsymbol{y}^a_1, \boldsymbol{y}^b_1)$) formula is as follows: 

\begin{equation}
\boldsymbol{y}^{new}_1 [m, j, k] = 
\begin{cases} 
\boldsymbol{y}^a_1[m, j, k],  \text{if } 0 \leq j < T, \\
\boldsymbol{y}^b_1[m, j- T, k],  \text{if } T \leq j < 2T.
\end{cases}
\end{equation}

Specifically, a new $\boldsymbol{y}^{new}_1$ is formed, with its elements in the dimensions $[B, 0-T, E]$ coming from $\boldsymbol{y}^a_1$, and the elements in the dimensions $[B, T-2T, E]$ coming from $y^b_1$. This formula represents $\boldsymbol{y}^{new}_1 = Cat(\boldsymbol{y}^a_1,\boldsymbol{y}^b_1)$ as stated in Algorithm \ref{Al1}, where $Cat(\cdot, \cdot)$ is concatenating $\boldsymbol{y}^a_1,\boldsymbol{y}^b_1$ along the patch dimension.

To enable the model to deeply fuse by using FAVBs multiple times, thereby allowing the feature representation model to effectively capture more scale information, we pass $\boldsymbol{y}^{new}_1$ through a larger ViT block $f_2(\cdot)$ to generate $\boldsymbol{y}^{newout}_1$. Then, $\boldsymbol{y}^{newout}_1$ is split into two features along the token dimension ${\boldsymbol{y}^a_2}$ and ${\boldsymbol{y}^b_2} \in \mathbb{R}^{B \times T \times E}$, and passed into the next layer. The specific splitting ($Spl(\boldsymbol{y}^{newout}_1)$) formula is as follows: 

\begin{equation}
\begin{cases}
\boldsymbol{y}^a_2[m, j, k] = \boldsymbol{y}^{newout}_1[m, j, k], \quad \text{if } 0 \leq j < T, \\
\boldsymbol{y}^b_2[m, j, k] = \boldsymbol{y}^{newout}_1[m, j+T, k], \quad \text{if } 0 \leq j < T.
\end{cases}
\end{equation}

Specifically, $y^{newout}_1$ is split into two new elements, $\boldsymbol{y}^a_2$ and $\boldsymbol{y}^b_2$, where $\boldsymbol{y}^a_2$ consists of the elements in dimensions $[B, 0:T, E]$ of $\boldsymbol{y}^{newout}_1$, and $\boldsymbol{y}^b_2$ consists of the elements in dimensions $[B, T:2T, E]$ of $\boldsymbol{y}^{newout}_1$. This formula represents  \( \boldsymbol{y}_{2}^a, \boldsymbol{y}_{2}^b = Spl(\boldsymbol{y}_1^{newout})\) as stated in Algorithm \ref{Al1}, where $Spl(\cdot)$ is splitting $\boldsymbol{y}^{newout}_i$ along the patch dimension.

This entire process is called Fusing-Augmenting ViT Blocks (FAVBs, Figure \ref{fig2}). The FAVBs is iterated $N$ times and normalized to obtain the final output features $\boldsymbol{y}^a_{N+1}$ and $\boldsymbol{y}^b_{N+1}$ \cite{ref19}. The final splitting manipulate is to project the two learned features into instance and cluster spaces for contrastive clustering.

For self-supervised deep clustering, a fusion strategy should preserve informative variance while facilitating effective similarity estimation. Our MFAVBs design meets this requirement by doubling the token dimension and thus retaining the full, unaltered representations from the two augmented branches\cite{ref55}; in contrast, element-wise addition can nullify discriminative cues when activations bear opposite signs. After concatenation, the ViT block leverages global self-attention to model complementarities: positional encodings let the network discern token provenance, so Query-Key interactions explicitly capture cross-view relations within a single layer. The subsequent split restores branch independence, allowing gradients to propagate separately and preventing premature homogenisation of representations—a frequent cause of cluster collapse in contrastive frameworks. Because no extra gating or cross-attention parameters are introduced\cite{ref56, ref57}, the method remains memory- and compute-efficient, which is especially valuable for small or noisy data sets where over-parameterisation leads to overfitting. Moreover, these more complex fusion strategies tend to prematurely homogenize the dual-branch information in contrastive learning, making it difficult to disentangle features during the subsequent decoupling process. Although such approaches may bring a certain degree of overall performance improvement, they are not well aligned with the architectural design of our current model. Instead, these methods are better suited for multimodal feature fusion frameworks that focus on enhancing a single unified feature representation. Conceptually, the pipeline mirrors the software Bridge Pattern: fusion forms the “bridge” that links implementation (feature extraction) and abstraction (instance- and cluster-level objectives), while the split decouples their optimisation. This structural separation enables fine-grained instance discrimination and global cluster coherence to evolve in tandem without mutual interference, ultimately yielding more robust and expressive clustering embeddings.

\subsection{Contrastive Clustering}\label{CC}

Contrastive learning aims to maximize the similarity between positive pairs while minimizing the similarity between negative pairs through self-supervised learning. In clustering tasks where there are no prior labels, pseudo-labels generated from data augmentation are used to construct positive and negative pairs at the instance level. SimCLR \cite{ref9} demonstrates that providing more negative samples in contrastive learning enhances feature extraction performance. Therefore, we consider the two augmentions of similar instances ${x_i^a, x_i^b}$ in a batch as a positive pair, while the remaining 2B - 2 samples ($\boldsymbol{x}^a \cup \boldsymbol{x}^b \setminus \{x_i^a, x_i^b\}$)are regarded as negative pairs.

To mitigate the information loss caused by contrastive loss, we adopt a similar approach to CC \cite{ref10} by not performing contrastive learning directly on the feature matrix. Instead, we project the feature matrix ${h^a_i,h^b_i}$ into an instance-level subspace using a two-layer nonlinear MLP $P_I(\cdot)$ to obtain $I^a_i, I^b_i = P_I(h^a_i), P_I(h^b_i)$. This approach not only increases the weight of important features but also suppresses unimportant or redundant features, thereby reducing information loss. Then, we adopt cosine similarity to evaluate the similarity between two samples, formulated as:

\begin{equation}
Sim(I_i^a, I_i^b) = \frac{(I_i^a)^T \cdot (I_i^b)}{\|I_i^a\| \|I_i^b\|}.
\end{equation}

Instance-level loss is computed through InfoNCE loss \cite{ref35}, which maximizes the ratio between the similarity of positive samples and the similarities of all samples. As a result, the model gradually enhances its ability to recognize positive samples while decreasing the similarity to negative samples during training.

\begin{equation}
\mathcal{L}_I^a(\boldsymbol{I}^a,\boldsymbol{I}^b) = - \frac{1}{B} {\sum_{i=1}^B \log \frac{\exp(Sim(I_i^a, I_i^b) / \tau)}{\sum_{j=1}^{2B-2} \exp(Sim(I_i^a, I_j) / \tau)}}.
\end{equation}

\begin{equation}
\mathcal{L}_{Cap} = -\mathbb{E}_{{x^t} \sim D}[\sum_{i=1}^B\log p(x^t_l|x^t_{<l},x^v)].
\end{equation}

Here, $\tau$ represents the temperature coefficient, used to regulate the model's ability to differentiate between negative samples. $I_j$ is the components of the instance-level negative sample set, that is $I_j \in \boldsymbol{I}^a \cap \boldsymbol{I}^b \setminus \{I_i^a, I_i^b\}$.

Simultaneously, we augment the instance-level contrastive loss with a cluster-level contrastive loss for joint optimization. The goal is to enhance the similarity between positive cluster pairs and diminish the similarity between negative cluster pairs. In a manner akin to the instance level, a softmax function is integrated into $P_I$ to provide soft labels for each clustering category $ C^a_i, C^b_i = P_C(h^a_i), P_C(h^b_i) $. The loss function remains based on InfoNCE, but we add an additional penalty function F to prevent most instances from being allocated to a single cluster:

\begin{equation}
\mathcal{L}_C^a(\boldsymbol{C}^a,\boldsymbol{C}^b) = - \frac{1}{B} {\sum_{i=1}^B \log \frac{\exp(Sim(C_i^a, C_i^b) / \tau)}{\sum_{j=1}^{2B-2} \exp(Sim(C_i^a, C_j) / \tau)}}.
\end{equation}

Here, $C_j$ is the components of the cluster-level negative sample set, that is $C_j \in \boldsymbol{C}^a \cap \boldsymbol{C}^b \setminus \{C_i^a, C_i^b\}$.

Ultimately, the total loss $\mathcal{L}$ is calculated by adding the instance-level and clustering-level loss values, and this total loss is used to update the parameters of each ViT block during the learning process. 

\begin{equation}
\mathcal{L} = \mathcal{L}_I^a(\boldsymbol{I}^a,\boldsymbol{I}^b) + \mathcal{L}_I^b(\boldsymbol{I}^b,\boldsymbol{I}^a) + \mathcal{L}_C^a(\boldsymbol{C}^a,\boldsymbol{C}^b) +\mathcal{L}_C^b(\boldsymbol{C}^b,\boldsymbol{C}^a)
\end{equation}

After the training is completed, the soft label of each sample is calculated, and the category with the highest probability is taken as the prediction result to achieve clustering.

\subsection{Parameter-free Interaction between CLIP Token and MFAVBs}\label{FMF}

Taking advantage of the strong zero-shot prior provided by CLIP~\cite{ref21}, we insert a CLIP-initialized, trainable multimodal anchor into the ViT input sequence. Concretely, we place the CLIP image-[CLS] embedding after the standard ViT [CLS] token so that it co-evolves with patch tokens through self-attention and residual paths. This multimodal anchor serves as a global semantic controller: it introduces class-level cues from the very beginning of training and persists across depth, encouraging the visual encoder to focus on category-relevant regions rather than relying on low-level similarities. \emph{Importantly, no new modules or parameters are added beyond this single token.} Rather than a plain concatenation, the anchor \emph{implicitly conditions} each MFAVB (Section~\ref{MFAVBs}) through mechanisms inherent to the backbone: (i) \textbf{softmax competition} in multi-head self-attention, where adding the CLIP key/value shifts attention mass of patch queries toward semantically aligned directions; (ii) \textbf{residual propagation} across depth, as the anchor aggregates evidence and influences subsequent layers; and (iii) \textbf{normalization coupling}, because LayerNorm operates over the concatenated sequence and thus re-centers token statistics toward class-relevant channels. Since the same anchor is shared across MFAVB branches under strong augmentations, it also provides a \textbf{cross-branch semantic reference} that implicitly aligns fused representations—all with negligible overhead (one extra token).

\begin{figure}[h] \centering \includegraphics[width=0.5\linewidth]{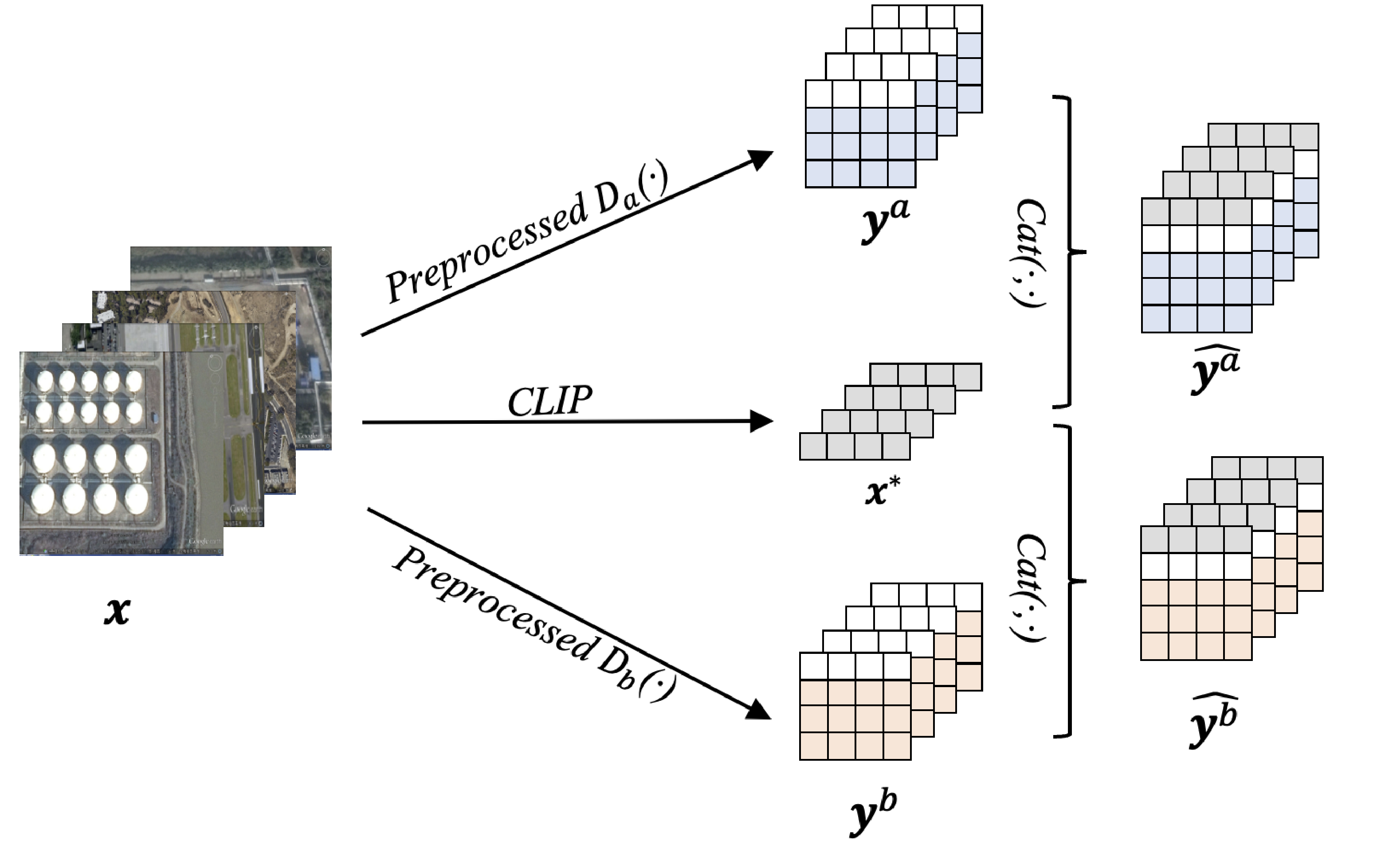} \caption{\small The fusion between original input data and CLIP features.} \label{fig6} \end{figure}

Given a batch of images $\boldsymbol{x}$, we obtain the CLIP image-[CLS] prior $c^0$ via the frozen CLIP encoder. Because CLIP aligns image--text modalities in a unified embedding space, no extra projection is required. We then insert $c^0$ into the zero position of the original input sequences $\boldsymbol{y}^{a}$ and $\boldsymbol{y}^{b}$ \emph{after} the standard ViT [CLS], producing augmented representations $\hat{\boldsymbol{y}}^{a}$ and $\hat{\boldsymbol{y}}^{b} \in \mathbb{R}^{B \times T \times E}$ with $T = P + 2$ tokens (patches $P$, ViT [CLS], and the multimodal anchor). As illustrated in Fig.~\ref{fig6}, this fusion seamlessly integrates multimodal priors into the ViT input without disrupting its visual structure, while enabling parameter-free conditioning of MFAVBs via shared attention, residual, and normalization pathways.

The enhanced input sequences $\hat{\boldsymbol{y}}^{a}$ and $\hat{\boldsymbol{y}}^{b}$ are subsequently fed into the MFAVBs module (Section \ref{MFAVBs}) to extract discriminative representations, followed by the contrastive clustering process described in Section \ref{CC}. Features are projected into instance-level and cluster-level subspaces to compute losses and update model parameters via backpropagation, ultimately yielding an optimized feature space for clustering.

For downstream classification or clustering, the sequence must be reduced to a single vector representation. Similar to BERT~\cite{ref45}, which uses the [CLS] token to represent the overall sentence meaning, we take the final-layer multimodal anchor (co-evolved CLIP token) as the global summary. This design ensures that the final representation integrates both visual and semantic cues from image--text pairs, thereby enhancing the model’s ability to distinguish visually similar yet semantically different samples and improving.

\subsection{Comparison}

In contrastive learning, the model learns representations by pulling the distance between positive sample pairs closer and pushing the distance between negative sample pairs farther apart. However, traditional contrastive learning methods often focus on learning features \cite{ref19} from the final output layer or deep layers, overlooking the potential of intermediate layer features. In contrast, by applying a fusion and separation strategy to the intermediate layer features in a two-branches network, contrastive learning can capture instance-level and cluster-level information at richer levels, resulting in more diverse features. This strategy not only helps the model integrate local features but also captures global features information at the cluster level, better reflecting the inherent diversity and complexity of the data.

Unlike some recent contrastive clustering methods, the flexible fusion and separation of intermediate layer features significantly enhance the representation capability and accelerate the training process. In these strategies, feature fusion explicitly strengthens the complementarity and similarity of information, preventing information loss, while feature separation helps maintain the uniqueness of each layer's features, enabling more precise multi-level representation. This fusion-separation strategy distinguishes itself from existing contrastive learning methods like CC \cite{ref10} and VTCC \cite{ref11}. The MFAVBs-CC method enhances semantic information exchange between intermediate layers through deep fusion and reduces information loss through multiple layers of integration. Additionally, the introduction of CLIP multimodal feature embedding further improves the model’s ability to distinguish between highly similar classes, enhancing its expressiveness.

Compared to direct addition or other fusion methods, the concatenation approach preserves the independence of each branch's features after connection without mixing them directly. This is particularly beneficial for the fusion of diverse or heterogeneous features, as it avoids potential information conflicts or mutual cancellation that can arise from direct addition. Furthermore, direct addition often treats the two branches as contributing equally, lacking the ability to model their importance or differences, which may hinder the full exploitation of the complementary nature of the branches.

Regarding the encoder, MFAVBs-CC adopts ViT encoders, which differs from SimCLR \cite{ref9} and CC \cite{ref10}. With its global self-attention mechanism, ViT can capture richer contextual information and long-range dependencies, making it particularly advantageous for handling remote dependencies between image elements, especially when dealing with long sequences. Therefore, MFAVBs-CC, using this feature representation encoder, achieves superior clustering performance, demonstrating stronger advantages over traditional CNN \cite{ref23} and ResNet \cite{ref17} in modeling global information.

\section{Experiments}

In this section, we present the clustering performance of MFAVBs-CC on four distinct remote sensing datasets, conduct a detailed analysis of the clustering results, and validate the model's reliability through experimental verification.

\begin{table}[h]
\centering
\caption{Seven datasets sample number and category number.}
\label{tab1}
\begin{tabular*}{0.75\linewidth}{@{\extracolsep\fill}lcc}
\toprule
Datasets & Samples & Clusters \\
\midrule
RSOD \cite{ref24} & 976 & 4  \\
UC-merced \cite{ref25} & 2,100 & 21  \\
SIRI-WHU \cite{ref26} & 2,400 & 12  \\
AID \cite{ref27} & 10,000 & 30  \\
ImageNet10 \cite{ref41}  & 13,000 & 10\\
CIFAR10 \cite{ref40} & 60,000 & 10 \\
CIFAR100 \cite{ref40} & 60,000 & 20 \\

\bottomrule
\end{tabular*}
\end{table}

\subsection{Datasets and Evaluation Metrics}

The datasets used in this paper include RSOD\cite{ref24}, UC-Merced\cite{ref25}, SIRI-WHU\cite{ref26}, AID\cite{ref27}, ImageNet10\cite{ref41}, CIFAR-10, and CIFAR-100\cite{ref40}. Among themBoth the training and test set are used for CIFAR10, CIFAR100, while only the training set is used for ImageNet10. The following table (Table \ref{tab1}) shows the number of samples and categories for each dataset.

To evaluate the performance of clustering, we used three widely-used evaluation metrics: normalized mutual information (NMI) \cite{ref36}, accuracy (ACC) \cite{ref37}, and adjusted marginal index (ARI) \cite{ref38}. During the data preprocessing, we utilized the identical conv stem as that of VTCC \cite{ref11} to transform the input image into a 1D sequence. We adopt the ViT-small \cite{ref28} model for training and use 8-layer ViT blocks, meaning the number N of FAVBs is 4. Considering that the dimensionality of the image features extracted by CLIP \cite{ref21} is [1, 512], we set the embedding dimension of vit-small to 512.  MFAVBs-CC underwent 500 training iterations with a stack size of 128. All experiments were performed on a V100-SXM2-32GB GPU with CUDA 11.0. The code has been uploaded to GitHub.\footnote{\url{https://github.com/sszhou-github/MFAVBs-CC}}

\subsection{Comparison of clustering performance with other methods}

In our study, due to space constraints, we compared our method with a traditional deep clustering method (K-means \cite{ref29}), two autoencoder clustering methods (Deep Embedded Clustering (DEC) \cite{ref30}, Improved Deep Embedding Clustering (IDFD) \cite{ref31}), deep clustering method CC \cite{ref10}, VTCC \cite{ref11}, and the two improved clustering methods based on ViT (MTDC \cite{ref39} and PDTE\cite{ref61}). For each base method, the optimal hyperparameters recommended in the respective original publications were used. The ACC, NMI, and ARI values for these seven methods across the seven datasets are shown in Table \ref{tab2} (Best results are in bold.).

\begin{table}[h]
\caption{The ACC, NMI and ARI scores of different image clustering methods on seven datasets.}\label{tab2}
\begin{tabular*}{\linewidth}{@{\extracolsep{\fill}}@{\hspace{-2pt}}l@{\hspace{-2pt}}c@{\hspace{-2pt}}c@{\hspace{-2pt}}c@{\hspace{-2pt}}c@{\hspace{-2pt}}c@{\hspace{-2pt}}c@{\hspace{-2pt}}c@{}}
\toprule
Method & RSOD & UC-Merced & SIRI-WHU & AID & ImageNet10 & CIFAR10 & CIFAR100\\
\midrule
\multicolumn{8}{c}{ACC}\\
kmeans\cite{ref29} & 0.388 & 0.200   & 0.229 & 0.163 & 0.241 & 0.229 & 0.130 \\
DEC\cite{ref30}   & 0.534 & 0.147 & 0.257 & 0.185 & 0.282 & 0.301 & 0.185 \\
IDFD\cite{ref31}  & 0.595 & 0.141 & 0.255 & 0.192 & 0.954 & 0.815 & 0.425 \\
CC\cite{ref10}    & 0.538 & 0.48  & 0.604 & 0.622 & 0.893 & 0.790  & 0.429 \\
VTCC\cite{ref11}  & 0.572 & 0.553 & 0.67  & 0.716 & 0.926 & 0.813 & 0.455 \\
MTDC\cite{ref39}  & 0.607 & 0.632 & 0.695 & 0.843 & 0.958 & 0.843 & 0.469 \\
SIC(direct)\cite{ref62} & 0.637 & 0.632 & 0.684 & 0.805 & 0.918 & 0.783 &0.459 \\
PDTE\cite{ref61}  & 0.679 & \textbf{0.688} & 0.820 & 0.852 & 0.960 & \textbf{0.903} & 0.509 \\
MFAVBs-CC & $\textbf{0.686} \scriptstyle \pm 0.1\%$ & $0.676 \scriptstyle \pm 0.2\%$ & $\textbf{0.743} \scriptstyle \pm 0.3\%$ & $\textbf{0.859} \scriptstyle \pm 0.9\%$ & $\textbf{0.962} \scriptstyle \pm 0.1\%$     & $0.883 \scriptstyle \pm 0.3\%$    & $\textbf{0.523}\scriptstyle \pm0.2\%$ \\

\midrule
\multicolumn{8}{c}{NMI}\\
kmeans\cite{ref29} & 0.162 & 0.204 & 0.145 & 0.209 & 0.119 & 0.087 & 0.084 \\
DEC\cite{ref30}   & 0.296 & 0.12  & 0.183 & 0.217 & 0.282 & 0.257 & 0.136 \\
IDFD\cite{ref31}  & 0.209 & 0.119 & 0.178 & 0.207 & 0.898 & 0.711 & 0.426 \\
CC\cite{ref10}    & 0.457 & 0.609 & 0.603 & 0.752 & 0.859 & 0.705 & 0.431 \\
VTCC\cite{ref11}  & 0.611 & 0.658 & 0.693 & 0.794 & 0.882     & 0.735     & 0.432 \\
MTDC\cite{ref39}   & 0.667 & 0.710  & 0.731 & 0.873 & 0.905 & 0.762 & 0.445 \\
SIC(direct)\cite{ref62} & 0.684 & 0.732 & 0.705 & 0.745 & 0.853 & 0.741 &0.369 \\
PDTE\cite{ref61}  & 0.687 & \textbf{0.764} & \textbf{0.787} & 0.882 & 0.905 & \textbf{0.842} & 0.489 \\
MFAVBs-CC & $\textbf{0.707}\scriptstyle \pm 0.3\%$ & $0.753 \scriptstyle \pm 1.2\%$ & $0.754 \scriptstyle \pm 0.4\%$ & $\textbf{0.895} \scriptstyle \pm 0.3\%$ & $\textbf{0.921} \scriptstyle \pm 0.2\%$    & $0.784 \scriptstyle \pm 0.5\%$     & $\textbf{0.496}\scriptstyle\pm 1.4\%$\\

\midrule
\multicolumn{8}{c}{ARI}\\
kmeans\cite{ref29} & 0.075 & 0.065 & 0.053 & 0.051 & 0.057 & 0.049 & 0.028 \\
DEC\cite{ref30}   & 0.325 & 0.053 & 0.083 & 0.075 & 0.203 & 0.161 & 0.05 \\
IDFD\cite{ref31}  & 0.362 & 0.042 & 0.079 & 0.073 & 0.901 & 0.663 & 0.264 \\
CC\cite{ref10}    & 0.371 & 0.356 & 0.45  & 0.55  & 0.822 & 0.637 & 0.266 \\
VTCC\cite{ref11}  & 0.482 & 0.453 & 0.554 & 0.622 & 0.891     & 0.664     & 0.281 \\
MTDC\cite{ref39}   & 0.566 & 0.531 & 0.593 & 0.781 & 0.903 & 0.722 & 0.297 \\
SIC(direct)\cite{ref62} & 0.537 & 0.552 & 0.604 & 0.795 & 0.908 & 0.743 &\textbf{0.539} \\
PDTE\cite{ref61}  & 0.598 & 0.578 & \textbf{0.699} & 0.791 & 0.919 & \textbf{0.817} & 0.319 \\
MFAVBs-CC & $\textbf{0.604}\scriptstyle \pm 0.7\%$ & $\textbf{0.587} \scriptstyle \pm 0.4\%$ & $0.658 \scriptstyle \pm 0.6\%$ & $\textbf{0.801} \scriptstyle \pm 0.1\%$ & $\textbf{0.928}\scriptstyle \pm 0.5\%$     & $0.759 \scriptstyle\pm 0.6\%$    & $0.327 \scriptstyle \pm 0.9\%$ \\

\bottomrule
\end{tabular*}
\end{table}

\begin{table}[h]
\caption{The ACC, NMI and ARI scores of different image clustering methods on seven datasets.}\label{tab2}
\begin{tabular*}{\linewidth}{@{\extracolsep{\fill}}@{\hspace{-2pt}}l@{\hspace{-2pt}}c@{\hspace{-2pt}}c@{\hspace{-2pt}}c@{\hspace{-2pt}}c@{\hspace{-2pt}}c@{\hspace{-2pt}}c@{\hspace{-2pt}}c@{}}
\toprule
Method & RSOD & UC-Merced & SIRI-WHU & AID & ImageNet10 & CIFAR10 & CIFAR100\\
\midrule
\multicolumn{8}{c}{ACC}\\
SIC(direct)\cite{ref62} & 0.637 & 0.632 & 0.684 & 0.805 & 0.918 & 0.783 &0.459 \\
MFAVBs-CC & $\textbf{0.686} \scriptstyle \pm 0.1\%$ & $\textbf{0.676} \scriptstyle \pm 0.2\%$ & $\textbf{0.743} \scriptstyle \pm 0.3\%$ & $\textbf{0.859} \scriptstyle \pm 0.9\%$ & $\textbf{0.962} \scriptstyle \pm 0.1\%$     & $\textbf{0.883} \scriptstyle \pm 0.3\%$    & $\textbf{0.523}\scriptstyle \pm0.2\%$ \\

\midrule
\multicolumn{8}{c}{NMI}\\

SIC(direct)\cite{ref62} & 0.684 & 0.732 & 0.705 & 0.745 & 0.853 & 0.741 &0.369 \\
MFAVBs-CC & $\textbf{0.707}\scriptstyle \pm 0.3\%$ & $\textbf{0.753} \scriptstyle \pm 1.2\%$ & $\textbf{0.754} \scriptstyle \pm 0.4\%$ & $\textbf{0.895} \scriptstyle \pm 0.3\%$ & $\textbf{0.921} \scriptstyle \pm 0.2\%$    & $\textbf{0.784} \scriptstyle \pm 0.5\%$     & $\textbf{0.496}\scriptstyle\pm 1.4\%$\\

\midrule
\multicolumn{8}{c}{ARI}\\
SIC(direct)\cite{ref62} & 0.537 & 0.552 & 0.604 & 0.795 & 0.908 & 0.743 &\textbf{0.539} \\
MFAVBs-CC & $\textbf{0.604}\scriptstyle \pm 0.7\%$ & $\textbf{0.587} \scriptstyle \pm 0.4\%$ & $\textbf{0.658} \scriptstyle \pm 0.6\%$ & $\textbf{0.801} \scriptstyle \pm 0.1\%$ & $\textbf{0.928}\scriptstyle \pm 0.5\%$     & $\textbf{0.759} \scriptstyle\pm 0.6\%$    & $\textbf{0.327} \scriptstyle \pm 0.9\%$ \\

\bottomrule
\end{tabular*}
\end{table}

Table \ref{tab2} results demonstrate that MFAVBs-CC not only shows improvements over the baseline methods presented in previous studies but also performs comparably to, and in some cases surpasses, advanced ViT-based contrastive clustering methods on certain datasets. Notably, ACC increased by an average of 0.087 across seven datasets compared to the baseline method VTCC \cite{ref11}, while NMI saw an average increase of 0.072 and ARI improved by 0.098. In particular, the accuracy in the RSOD, UC-Merced, and AID datasets improved by more than 0.1. These datasets comprise remote sensing images that contain numerous satellite images. In these datasets, overlapping images are common; for example, an airport image may include a playground. Both VTCC and CC have limited capacity to distinguish such images, which can result in misclassifications. Our method incorporates multimodal data to enhance the model with high-level semantic information. Furthermore, the Concatenating-Splitting ViT Blocks allows the model to address both instance and class information simultaneously, effectively alleviating misclassification and improving clustering performance.

\subsection{Ablation Experiments} 

We performed the following ablation experiments to explore the effectiveness of the two new modules in MFAVBs-CC: MFAVBs and the fusion of CLIP features with the input data. 

$\bullet$ MethodA: To examine whether the MFAVBs module enhances clustering accuracy through explicit interaction, we replace the input data of MFAVBs-CC with the original preprocessed augmentions, excluding the features extracted from CLIP.

$\bullet$ MethodB: To explore whether the fusion of the original preprocessed augmentions and features extracted from CLIP can guide the model to better learn image features, we replace the encoder backbone of MFAVBs-CC with a ViT encoder, which is the same as the backbone used in VTCC, consisting of 8 layers of ViT blocks. 

Experiments were performed across seven datasets, yielding the results shown in Table \ref{tab3} (Best results are in bold.).

\begin{table}[h]
\caption{The ACC, NMI and ARI scores of four image clustering methods on seven datasets. }\label{tab3}
\begin{tabular*}{0.85\linewidth}{@{\extracolsep{\fill}}@{\hspace{-2pt}}l@{\hspace{-2pt}}c@{\hspace{-2pt}}c@{\hspace{-2pt}}c@{\hspace{-2pt}}c@{\hspace{-2pt}}c@{\hspace{-2pt}}c@{\hspace{-2pt}}c@{}}
\toprule
Method & RSOD & UC-Merced & SIRI-WHU & AID & ImageNet10 & CIFAR10 & CIFAR100\\
\midrule
\multicolumn{8}{c}{ACC}\\
VTCC\cite{ref11}  & 0.572 & 0.553 & 0.670  & 0.716 & 0.926     & 0.813     & 0.455 \\
MethodA & 0.645 & 0.652 & 0.719 & 0.824 & 0.953     & 0.872    & 0.513 \\
MethodB & 0.667 & 0.661 & 0.731 & 0.839 & 0.947     & 0.867     & 0.491 \\
MFAVBs-CC & \textbf{0.686} & \textbf{0.676} & \textbf{0.743} & \textbf{0.859} & \textbf{0.962}     & \textbf{0.883}     & \textbf{0.523} \\
\midrule
\multicolumn{8}{c}{NMI}\\
VTCC\cite{ref11}  & 0.611 & 0.658 & 0.693 & 0.794 & 0.882     & 0.735     & 0.432 \\
MethodA & 0.673 & 0.732 & 0.746 & 0.864 & 0.919     & 0.771     & 0.486 \\
MethodB & 0.669 & 0.741 & 0.753 & 0.852 & 0.914     & 0.780     & 0.472 \\
MFAVBs-CC & \textbf{0.707} & \textbf{0.753} & \textbf{0.754} & \textbf{0.895} & \textbf{0.921}     & \textbf{0.784}     & \textbf{0.496} \\
\midrule
\multicolumn{8}{c}{ARI}\\
VTCC\cite{ref11}  & 0.482 & 0.453 & 0.554 & 0.622 & 0.891     & 0.664     & 0.281 \\
MethodA & 0.581 & 0.552 & 0.604 & 0.743 & 0.915     & 0.747     & 0.305 \\
MethodB & 0.593 & 0.566 & 0.619 & 0.775 & 0.917     & 0.752     & 0.311 \\
MFAVBs-CC & \textbf{0.604} & \textbf{0.587} & \textbf{0.628} & \textbf{0.801} & \textbf{0.928}     & \textbf{0.759}     & \textbf{0.327} \\
\bottomrule
\end{tabular*}
\end{table}

Compared to VTCC, which uses only parameter sharing for implicit interaction, MethodA employs an explicit Concatenating-Splitting framework that allows the baseline contrastive learning network to fully utilize both instance-level and cluster-level information for higher-quality feature extraction. As shown in Table \ref{tab3}, the new MFAVBs framework improves clustering accuracy by an average of 0.067 over VTCC. This Concatenating-Splitting framework could also be applied to other similar dual-branch contrastive learning networks with shared parameters. By fusing input data with CLIP features, Method B guides the model to better differentiate between similar images. Compared to VTCC, clustering accuracy increases by an average of 0.097, especially on four remote sensing datasets. This CLS token-style fusion effectively combines CLIP features with image features, introducing subtle semantic distinctions into the representations of similar images. MFAVBs-CC leverages the benefits of both methods, further enhancing clustering performance and demonstrating strong generalization across seven datasets.

\subsection{Visualization Analysis of FAVBs}

\begin{figure}[h]
    \centering
    \includegraphics[width=0.75\linewidth]{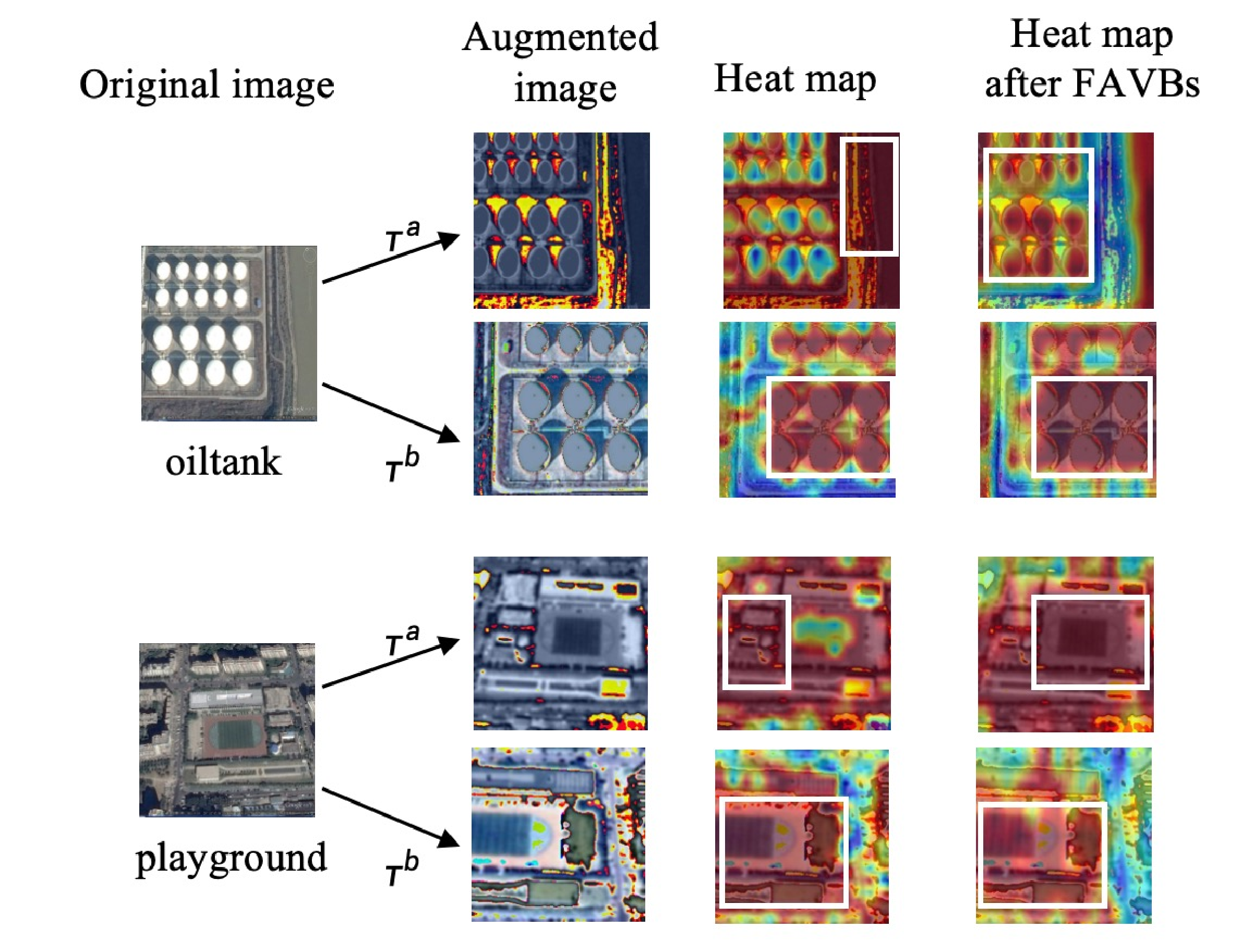}
    \caption{\small Self-attention heat maps before and after fusion.  The red regions of the heat map indicate higher self-attention weights, while the blue regions indicate lower self-attention weights.}
    \label{fig3}
\end{figure}

To investigate the complementarity and similarity between two augmentions, we visualize the self-attention matrix.  We randomly select two images from the RSOD dataset and apply two augmented methods to obtain augmented images. Subsequently, the trained MFAVBs-CC model is utilized to process the two augmented images, and the weight parameters of two shared-weight ViTs and concatenating larger ViT in the final FAVBs are extracted in order to obtain their respective self-attention weight heat maps. The comparative visualization of two images from the RSOD dataset before and after fusion is shown in Figure \ref{fig3}.

In the first oiltank image of Fig. \ref{fig3}, due to different sizes of the random crop, $\mathcal{T}^a$ results in a complete four-row oiltank, while $\mathcal{T}^b$ results in a three-row oiltank. $\mathcal{T}^a$ can compensate for missing features in $\mathcal{T}^b$. During grayscale processing, $\mathcal{T}^a$ makes it difficult to identify oiltank, focusing more on the road after ViT. In contrast, $\mathcal{T}^b$ directs more attention to the oiltank after ViT. After fusing the two images, attention in the grayscale image is focused entirely on the oiltank. Therefore, two augmentation exhibit a complementary relationship. The results demonstrate that our fusion technique effectively exploits this complementary and similarity relationship, and improves model feature representation.

\subsection{Time Complexity Analysis}

The MFVBs module inevitably increases both computation and memory costs by feeding ViT with longer token sequences. In a standard ViT block, the self-attention mechanism dominates time complexity, scaling as $O(T^2E)$, where $T$ is sequence length and $E$ is feature dimensionality. In our baseline VTCC model, each of two branches generates $T=198$ tokens of dimension $E=512$ and independently passes them through eight blocks (16 attentions in total). By contrast, MFAVBs concatenates the two branches’ features at layers 2, 4, 6, and 8—doubling the sequence length to $2T=396$ in four blocks while leaving the other four unchanged. The resulting unit complexity is a 25\% increase over the baseline. Crucially, however, this fusion accelerates convergence: our model reaches comparable or superior clustering quality in only 60\% of the original training epochs. Thus, the modest per-epoch slowdown is offset by a 20\%-30\% reduction in total training time, making the trade-off favorable. On the memory side, assume batch size $B=128$, $h=8$ heads, and 4-byte floats. Across four such layers, the peak memory footprint increases by roughly 1.8 GB—well within the 32 GB of a V100-SXM2. The calculation comparison is shown in the following table \ref{tab4}.

\begin{table}[htbp]
\centering
\caption{Time complexity and Memory usage comparison table}\label{tab4}
\label{tab:dataset_comparison}
\begin{tabular*}{0.85\linewidth}{@{\extracolsep{\fill}}@{\hspace{-2pt}}l@{\hspace{-2pt}}c@{\hspace{-2pt}}c@{}}
\toprule
& Time complexity & Memory\\
\midrule
VTCC & $16 \times 198^2 \times 512 $ & $4 \times 198^2 \times 512 \;+\; 4 \times 396^2 \times 512$\\
MFAVBs-CC & $128 \times 8 \times 396^2 \times 4~\mathrm{bytes}$& $128 \times 8 \times 198^2 \times 4~\mathrm{bytes}$\\
\bottomrule
\end{tabular*}
\vspace{5pt}
\footnotesize
\end{table}

The comparison of the time of a single epoch of VTCC in different data sets is shown in the following table \ref{tab5}. In addition, we calculated the total training time for our model. We set a stopping tolerance of 20 epochs, meaning that if the total loss did not decrease within 20 consecutive epochs, the model was considered converged. The table \ref{tab5} presents the total training time of our model compared with the baseline model across all datasets.

\begin{table}[htbp]
\centering
\caption{Average time spent on single and total epoch training time}\label{tab5}
\label{tab:dataset_comparison}
\begin{tabular*}{\linewidth}{@{\extracolsep{\fill}}@{\hspace{-2pt}}l@{\hspace{-2pt}}c@{\hspace{-2pt}}c@{\hspace{-2pt}}c@{\hspace{-2pt}}c@{\hspace{-2pt}}c@{\hspace{-2pt}}c@{\hspace{-2pt}}c@{}}
\toprule
& RSOD & UC-Merced & SIRI-WHU & AID & ImageNet10 & CIFAR10 & CIFAR100\\
\midrule
VTCC(single/s) & 6.383 & 12.470 & 13.481 & 64.668 & 75.148 & 371.033 & 371.369\\
MFAVBs-CC(single/s) & 8.363 & 15.203& 15.772 & 70.068 & 82.005 & 387.568 & 387.783 \\
VTCC(total/min) & 34.042 & 83.133 & 89.873 & 700.57 & 826.628 & 6183.883 & 6189.483\\
MFAVBs-CC(total/min) & 27.877 & 76.015 & 78.86 & 618.934  & 724.378 & 4198.653  & 4200.983 \\
\bottomrule
\end{tabular*}
\vspace{5pt}
\footnotesize
\end{table}

As shown in the following Fig \ref{tab5}, per-epoch run-times on all datasets rose only marginally, while total training wall-clock time decreased substantially. MFAVBs thus delivers faster deep contrastive clustering convergence with only modest additional compute and memory overhead.

\subsection{Sensitivity Analysis}

In this section, we analyze the impact of the core modules of MFAVBs-CC on clustering accuracy through experiments, including the number of intermediate layers and the number of FAVBs on four remote sensing datasets.

In our MFAVBs framework, we employ two distinct data augmentation pipelines at the input level to enhance the model's robustness and improve clustering performance.The first augmentation pipeline includes: random cropping with a scale range of 8\% to 100\%, ColorJitter with an 80\% probability, RandomGrayscale with a 20\% probability, Gaussian blur, and RandomHorizontalFlip.The second pipeline builds upon the first by additionally incorporating a Solarize transformation with a 20\% probability.These augmentions are designed not only to enrich the diversity of the input data but also to introduce controlled noise, thereby improving the model’s resilience to variations in visual appearance.

\subsubsection{Influence of the number of the Intermediate Layer}

\begin{figure}[h]
    \centering
    \includegraphics[width=0.5\linewidth]{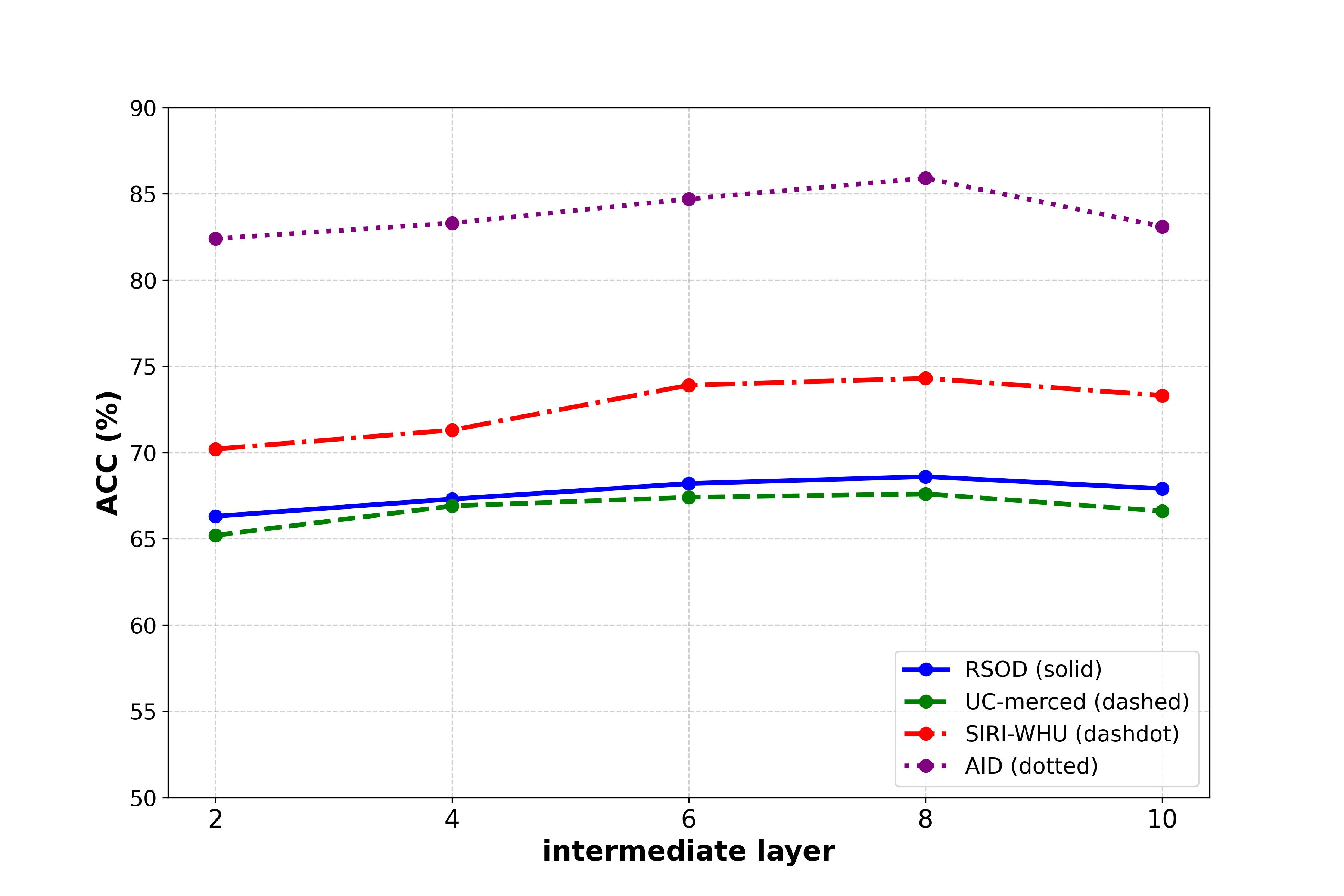}
    \caption{\small Line graph of influence of the number of intermediate layers on four datasets.}
    \label{fig4}
\end{figure}

The number of intermediate layers in the encoder was set to 2, 4, 6, 8, and 10, with one FAVBs module used every two layers. For example, with 2 intermediate layers, one FAVBs is used; with 4 layers, two FAVBs are used, and so on. The clustering accuracy variation curve is presented in Figure \ref{fig4} (only the ACC changes are shown here).

As shown in Figure \ref{fig4}, clustering accuracy increases approximately linearly with the number of intermediate layers. The highest clustering accuracy is achieved when the number of intermediate layers reaches 8. However, as the number of layers continues to increase, clustering accuracy begins to decline, with a noticeable drop at 10 layers. This trend mirrors the results observed in the experiments with VTCC \cite{ref11} and Deep Fusion \cite{ref19}, which could be attributed to overfitting due to the limited dataset size. In particular, the RSOD dataset, with only 976 images, shows the most significant drop in clustering accuracy when the number of intermediate layers reaches 10, as indicated in Figure \ref{fig4}. Therefore, the experiment suggests that the most suitable configuration is using ViT-small with 8 intermediate layers, which strikes a balance between clustering accuracy and efficiency.

\subsubsection{Influence of the number of FAVBs}

\begin{figure}[h]
    \centering
    \includegraphics[width=0.5\linewidth]{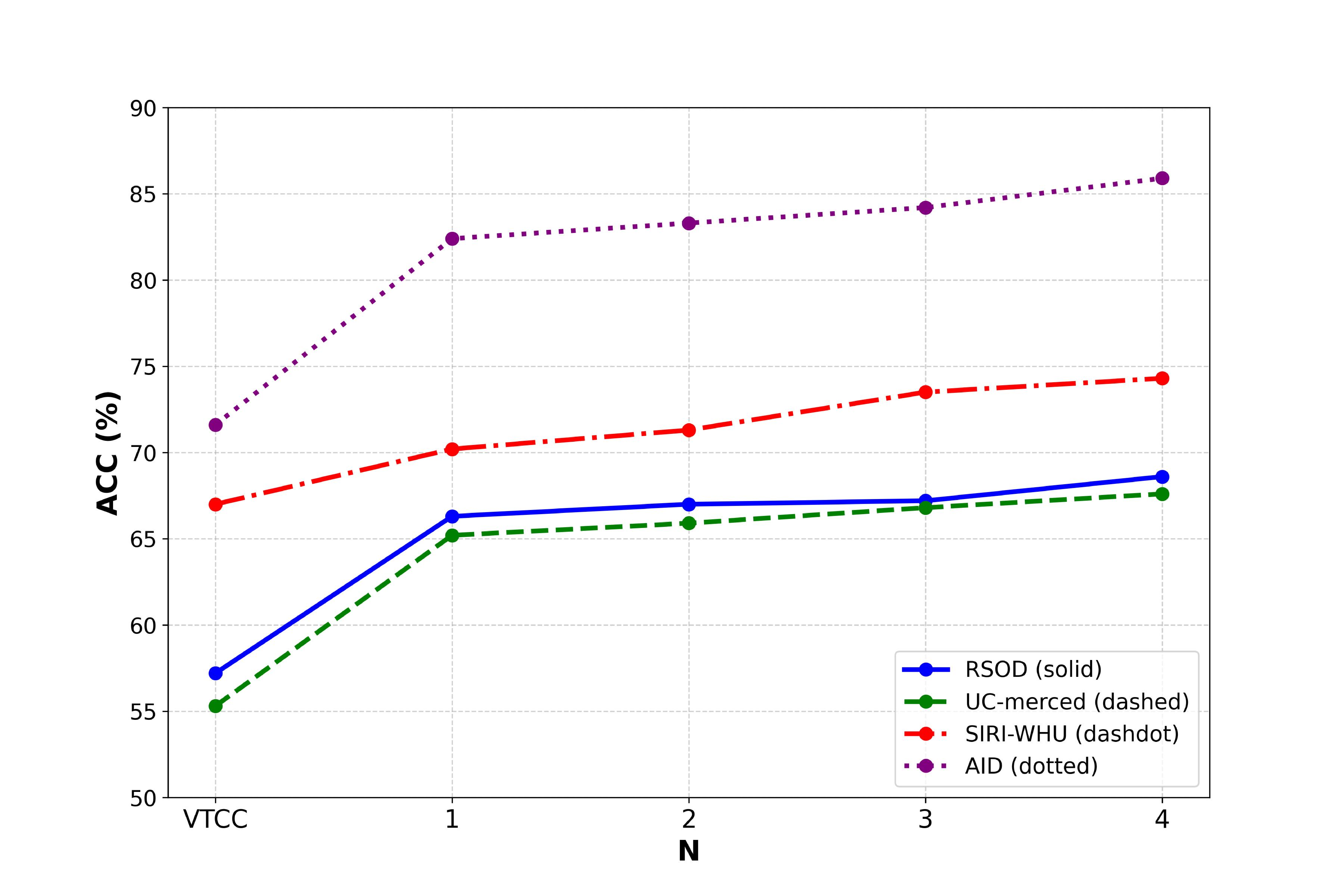}
    \caption{\small Line graph of influence of the number of N on four datasets.}
    \label{fig5}
\end{figure}

To investigate the impact of fusion times on clustering effectiveness, we conducted experiments on four datasets using different numbers of FAVBs iterations ($N$) when the number of intermediate layers was 8. Due to the numerous permutations and combinations of interactions, for simplicity, FAVBs is only applied to the initial $2N$ layers. For instance, when N=1, the first two layers utilize a FAVBs while the subsequent six layers do not undergo fusion; when N=2, the first four layers utilize two FAVBs while the subsequent four layers do not undergo fusion, and so on. When feature fusion is absent (i.e., $N=0$), it serves as the baseline method VTCC. The clustering accuracy variation curve is presented in Figure \ref{fig5} (only the ACC changes are shown here). 

As depicted in Fig \ref{fig5}, the results indicate that the FAVBs fusion technology enhances clustering performance, and when $N=4$, MFAVBs-CC achieves optimal clustering performance. This demonstrates that multiple FAVBs can progressively integrate diverse levels of information, thereby accumulating and reinforcing global contextual information. By stacking multiple FAVBs, the model can repeatedly capture and refine unique and complementary features between instance-level and cluster-level representations, leveraging these interactions to extract superior image features. In future work, with sufficient GPU performance and memory, employing a greater number of intermediate layers and FAVBs when dealing with larger datasets may lead to further improvements in clustering accuracy.

\subsection{Robustness Analysis}
We conducted experiments on the RSOD, UC-Merced, SIRI-WHU and AID datasets by applying random masking to simulate real noisy data. Specifically, before performing data augmentation, 30\%–50\% of each image area was masked out (blackened) to generate a certain amount of realistic noise-corrupted samples. The model was then retrained from scratch, and the results are summarized in the Table \ref{tab6}.

\begin{table}[htbp]
\centering
\caption{Comparison of model effects before and after adding noise}\label{tab6}
\label{tab:dataset_comparison}
\begin{tabular*}{0.85\linewidth}{@{\extracolsep{\fill}}@{\hspace{-2pt}}l@{\hspace{-2pt}}c@{\hspace{-2pt}}c@{\hspace{-2pt}}c@{\hspace{-2pt}}c@{}}
\toprule
& RSOD & UC-Merced & SIRI-WHU & AID \\
\midrule
MFAVBs-CC & 0.686 & 0.676 & 0.743 & 0.859 \\
add noise & 0.683 & 0.672 & 0.742 & 0.851 \\
\bottomrule
\end{tabular*}
\vspace{5pt}
\footnotesize
\end{table}

As shown in the table, the performance of MFAVBs-CC on these newly generated noisy datasets decreased by approximately 1\% compared to the original datasets, demonstrating that the model maintains strong clustering performance and robustness seven when faced with complex noisy data conditions.

\section{Conclusion}

In this paper, MFAVBs-CC leverages multiple fusion of intermediate features to enhance the complementarity and similarity between two augmentions and utilizes the MFAVBs framework to improve intermediate semantic linkage. The encoder of MFAVBs can be adapted to various application scenarios by replacing other feature dimensionality retention models. Given sufficient hardware resources, scaling up the number of intermediate layers and frequency of fusion can effectively address large-scale data aggregation tasks.

\section{Acknowledgments}

This work is supported by the National Natural Science Foundation of China (Grants numbers: 12571568), Shaanxi Fundamental Science Research Project for Mathematics and Physics (Grant No. 23JSY039) and the Natural Science Foundation of Shanxi Province, China (2024JCYBMS-051).

\section{Data Availability} 
Some or all data, models, or code generated or used during the study are available from the corresponding author by request.
\section{Declarations}
Conflict of interest/Competing interests 

The authors declare that they have no known competing financial interests or personal relationships that could have appeared to influence the work reported in this paper.



\begin{thebibliography}{44}
\ifx \bisbn   \undefined \def \bisbn  #1{ISBN #1}\fi
\ifx \binits  \undefined \def \binits#1{#1}\fi
\ifx \bauthor  \undefined \def \bauthor#1{#1}\fi
\ifx \batitle  \undefined \def \batitle#1{#1}\fi
\ifx \bjtitle  \undefined \def \bjtitle#1{#1}\fi
\ifx \bvolume  \undefined \def \bvolume#1{\textbf{#1}}\fi
\ifx \byear  \undefined \def \byear#1{#1}\fi
\ifx \bissue  \undefined \def \bissue#1{#1}\fi
\ifx \bfpage  \undefined \def \bfpage#1{#1}\fi
\ifx \blpage  \undefined \def \blpage #1{#1}\fi
\ifx \burl  \undefined \def \burl#1{\textsf{#1}}\fi
\ifx \doiurl  \undefined \def \doiurl#1{\url{https://doi.org/#1}}\fi
\ifx \betal  \undefined \def \betal{\textit{et al.}}\fi
\ifx \binstitute  \undefined \def \binstitute#1{#1}\fi
\ifx \binstitutionaled  \undefined \def \binstitutionaled#1{#1}\fi
\ifx \bctitle  \undefined \def \bctitle#1{#1}\fi
\ifx \beditor  \undefined \def \beditor#1{#1}\fi
\ifx \bpublisher  \undefined \def \bpublisher#1{#1}\fi
\ifx \bbtitle  \undefined \def \bbtitle#1{#1}\fi
\ifx \bedition  \undefined \def \bedition#1{#1}\fi
\ifx \bseriesno  \undefined \def \bseriesno#1{#1}\fi
\ifx \blocation  \undefined \def \blocation#1{#1}\fi
\ifx \bsertitle  \undefined \def \bsertitle#1{#1}\fi
\ifx \bsnm \undefined \def \bsnm#1{#1}\fi
\ifx \bsuffix \undefined \def \bsuffix#1{#1}\fi
\ifx \bparticle \undefined \def \bparticle#1{#1}\fi
\ifx \barticle \undefined \def \barticle#1{#1}\fi
\bibcommenthead
\ifx \bconfdate \undefined \def \bconfdate #1{#1}\fi
\ifx \botherref \undefined \def \botherref #1{#1}\fi
\ifx \bchapter \undefined \def \bchapter#1{#1}\fi
\ifx \bbook \undefined \def \bbook#1{#1}\fi
\ifx \bcomment \undefined \def \bcomment#1{#1}\fi
\ifx \oauthor \undefined \def \oauthor#1{#1}\fi
\ifx \citeauthoryear \undefined \def \citeauthoryear#1{#1}\fi
\ifx \endbibitem  \undefined \def \endbibitem {}\fi
\ifx \bconflocation  \undefined \def \bconflocation#1{#1}\fi
\ifx \arxivurl  \undefined \def \arxivurl#1{\textsf{#1}}\fi
\csname PreBibitemsHook\endcsname

\bibitem[\protect\citeauthoryear{Liu}{2021}]{ref34}
\begin{botherref}
\oauthor{\bsnm{Liu}, \binits{R.}}:
Understand and improve contrastive learning methods for visual representation: A review.
arXiv:2106.03259
(2021)
\end{botherref}
\endbibitem

\bibitem[\protect\citeauthoryear{MacQueen et~al.}{1967}]{ref29}
\begin{barticle}
\bauthor{\bsnm{MacQueen}, \binits{J.}}, \betal:
\batitle{Some methods for classification and analysis of multivariate observations}.
\bjtitle{Proceedings of the fifth Berkeley symposium on mathematical statistics and probability}
\bvolume{1}(\bissue{14}),
\bfpage{281}--\blpage{297}
(\byear{1967}).
\bcomment{Oakland, CA, USA}
\end{barticle}
\endbibitem

\bibitem[\protect\citeauthoryear{Bezdek et~al.}{1984}]{ref53}
\begin{barticle}
\bauthor{\bsnm{Bezdek}, \binits{J.C.}},
\bauthor{\bsnm{Ehrlich}, \binits{R.}},
\bauthor{\bsnm{Full}, \binits{W.}}:
\batitle{Fcm: The fuzzy c-means clustering algorithm}.
\bjtitle{Computers \& Geosciences}
\bvolume{10}(\bissue{2}),
\bfpage{191}--\blpage{203}
(\byear{1984})
\doiurl{10.1016/0098-3004(84)90020-7}
\end{barticle}
\endbibitem

\bibitem[\protect\citeauthoryear{Zhou et~al.}{2021}]{ref50}
\begin{barticle}
\bauthor{\bsnm{Zhou}, \binits{S.}},
\bauthor{\bsnm{Li}, \binits{D.}},
\bauthor{\bsnm{Zhang}, \binits{Z.}},
\bauthor{\bsnm{Ping}, \binits{R.}}:
\batitle{A new membership scaling fuzzy c-means clustering algorithm}.
\bjtitle{IEEE Transactions on Fuzzy Systems}
\bvolume{29}(\bissue{9}),
\bfpage{2810}--\blpage{2818}
(\byear{2021})
\doiurl{10.1109/TFUZZ.2020.3003441}
\end{barticle}
\endbibitem

\bibitem[\protect\citeauthoryear{Li et~al.}{2023}]{ref51}
\begin{barticle}
\bauthor{\bsnm{Li}, \binits{D.}},
\bauthor{\bsnm{Zhou}, \binits{S.}},
\bauthor{\bsnm{Pedrycz}, \binits{W.}}:
\batitle{Accelerated fuzzy c-means clustering based on new affinity filtering and membership scaling}.
\bjtitle{IEEE Transactions on Knowledge and Data Engineering}
\bvolume{35}(\bissue{12}),
\bfpage{12337}--\blpage{12349}
(\byear{2023})
\doiurl{10.1109/TKDE.2023.3273274}
\end{barticle}
\endbibitem

\bibitem[\protect\citeauthoryear{Li et~al.}{2024}]{ref52}
\begin{barticle}
\bauthor{\bsnm{Li}, \binits{D.}},
\bauthor{\bsnm{Zhou}, \binits{S.}},
\bauthor{\bsnm{Zeng}, \binits{T.}},
\bauthor{\bsnm{Chan}, \binits{R.H.}}:
\batitle{Multi-prototypes convex merging based k-means clustering algorithm}.
\bjtitle{IEEE Transactions on Knowledge and Data Engineering}
\bvolume{36}(\bissue{11}),
\bfpage{6653}--\blpage{6666}
(\byear{2024})
\doiurl{10.1109/TKDE.2023.3342209}
\end{barticle}
\endbibitem

\bibitem[\protect\citeauthoryear{Caron et~al.}{2020}]{ref2}
\begin{barticle}
\bauthor{\bsnm{Caron}, \binits{M.}},
\bauthor{\bsnm{Misra}, \binits{I.}},
\bauthor{\bsnm{Mairal}, \binits{J.}},
\bauthor{\bsnm{Goyal}, \binits{P.}},
\bauthor{\bsnm{Bojanowski}, \binits{P.}},
\bauthor{\bsnm{Joulin}, \binits{A.}}:
\batitle{Unsupervised learning of visual features by contrasting cluster assignments}.
\bjtitle{Advances in neural information processing systems}
\bvolume{33},
\bfpage{9912}--\blpage{9924}
(\byear{2020})
\end{barticle}
\endbibitem

\bibitem[\protect\citeauthoryear{Reimers and Gurevych}{2019}]{ref3}
\begin{bchapter}
\bauthor{\bsnm{Reimers}, \binits{N.}},
\bauthor{\bsnm{Gurevych}, \binits{I.}}:
\bctitle{Sentence-bert: Sentence embeddings using siamese bert-networks}.
In: \bbtitle{Proceedings of the 2019 Conference on Empirical Methods in Natural Language Processing and the 9th International Joint Conference on Natural Language Processing (EMNLP-IJCNLP)},
pp. \bfpage{3982}--\blpage{3992}.
\bpublisher{Association for Computational Linguistics},
\blocation{Hong Kong, China}
(\byear{2019})
\end{bchapter}
\endbibitem

\bibitem[\protect\citeauthoryear{Yang et~al.}{2024}]{ref4}
\begin{barticle}
\bauthor{\bsnm{Yang}, \binits{S.}},
\bauthor{\bsnm{Cui}, \binits{L.}},
\bauthor{\bsnm{Wang}, \binits{L.}},
\bauthor{\bsnm{Wang}, \binits{T.}}:
\batitle{Cross-modal contrastive learning for multimodal sentiment recognition}.
\bjtitle{Applied Intelligence}
\bvolume{154},
\bfpage{4260}--\blpage{4276}
(\byear{2024})
\doiurl{10.1007/s10489-024-05355-8}
\end{barticle}
\endbibitem

\bibitem[\protect\citeauthoryear{Ren et~al.}{2023}]{ref46}
\begin{barticle}
\bauthor{\bsnm{Ren}, \binits{H.}},
\bauthor{\bsnm{Zheng}, \binits{Z.}},
\bauthor{\bsnm{Wu}, \binits{Y.}},
\bauthor{\bsnm{Lu}, \binits{H.}}:
\batitle{Daco: domain-agnostic contrastive learning for visual place recognition}.
\bjtitle{Applied Intelligence}
\bvolume{53},
\bfpage{21827}--\blpage{21840}
(\byear{2023})
\doiurl{10.1007/s10489-023-04629-x}
\end{barticle}
\endbibitem

\bibitem[\protect\citeauthoryear{Wang et~al.}{2024}]{ref65}
\begin{bchapter}
\bauthor{\bsnm{Wang}, \binits{H.}},
\bauthor{\bsnm{Zhou}, \binits{W.}},
\bauthor{\bsnm{Wen}, \binits{J.}},
\bauthor{\bsnm{Qiao}, \binits{S.}}:
\bctitle{Multiple hypergraph convolutional network social recommendation using dual contrastive learning}.
In: \bbtitle{Data Mining and Knowledge Discovery},
\beditor{\bparticle{Editors: }\bauthor{\bsnm{Data Mining}}, \binits{H.}},
(\byear{2024})
\bfpage{1929}--\blpage{1957}.
\doiurl{10.1007/s10618-024-01021-2}.
\end{bchapter}
\endbibitem


\bibitem[\protect\citeauthoryear{Chopra et~al.}{2005}]{ref1}
\begin{bchapter}
\bauthor{\bsnm{Chopra}, \binits{S.}},
\bauthor{\bsnm{Hadsell}, \binits{R.}},
\bauthor{\bsnm{LeCun}, \binits{Y.}}:
\bctitle{Learning a similarity metric discriminatively, with application to face verification}.
In: \bbtitle{2005 IEEE Computer Society Conference on Computer Vision and Pattern Recognition (CVPR'05)},
vol. \bseriesno{1}.
\bconflocation{San Diego, America},
pp. \bfpage{539}--\blpage{5461}
(\byear{2005}).
\doiurl{10.1109/CVPR.2005.202}
\end{bchapter}
\endbibitem

\bibitem[\protect\citeauthoryear{Wu et~al.}{2018}]{ref8}
\begin{bchapter}
\bauthor{\bsnm{Wu}, \binits{Z.}},
\bauthor{\bsnm{Xiong}, \binits{Y.}},
\bauthor{\bsnm{Yu}, \binits{S.X.}},
\bauthor{\bsnm{Lin}, \binits{D.}}:
\bctitle{Unsupervised feature learning via non-parametric instance discrimination}.
In: \bbtitle{Proceedings of the IEEE Conference on Computer Vision and Pattern Recognition (CVPR)},
\bconflocation{Salt Lake City, USA}
(\byear{2018}).
\doiurl{10.1109/CVPR.2018.00393}
\end{bchapter}
\endbibitem

\bibitem[\protect\citeauthoryear{Chen et~al.}{2023}]{ref33}
\begin{barticle}
\bauthor{\bsnm{Chen}, \binits{Z.}},
\bauthor{\bsnm{Lin}, \binits{K.-Y.}},
\bauthor{\bsnm{Zheng}, \binits{W.-S.}}:
\batitle{Consistent intra-video contrastive learning with asynchronous long-term memory bank}.
\bjtitle{IEEE Transactions on Circuits and Systems for Video Technology}
\bvolume{33}(\bissue{3}),
\bfpage{1168}--\blpage{1180}
(\byear{2023})
\doiurl{10.1109/TCSVT.2022.3207174}
\end{barticle}
\endbibitem

\bibitem[\protect\citeauthoryear{Chen et~al.}{2020}]{ref9}
\begin{barticle}
\bauthor{\bsnm{Chen}, \binits{T.}},
\bauthor{\bsnm{Kornblith}, \binits{S.}},
\bauthor{\bsnm{Norouzi}, \binits{M.}},
\bauthor{\bsnm{Hinton}, \binits{G.}}:
\batitle{A simple framework for contrastive learning of visual representations}.
\bjtitle{Proceedings of the 37th International Conference on Machine Learning, PMLR}
\bvolume{119},
\bfpage{1597}--\blpage{1607}
(\byear{2020})
\end{barticle}
\endbibitem

\bibitem[\protect\citeauthoryear{Li et~al.}{2021}]{ref10}
\begin{bchapter}
\bauthor{\bsnm{Li}, \binits{Y.}},
\bauthor{\bsnm{Hu}, \binits{P.}},
\bauthor{\bsnm{Liu}, \binits{Z.}},
\bauthor{\bsnm{Peng}, \binits{D.}},
\bauthor{\bsnm{Zhou}, \binits{J.T.}},
\bauthor{\bsnm{Peng}, \binits{X.}}:
\bctitle{Contrastive clustering}.
In: \bbtitle{Proceedings of the AAAI Conference on Artificial Intelligence},
vol. \bseriesno{35},
pp. \bfpage{8547}--\blpage{8555}
(\byear{2021})
\end{bchapter}
\endbibitem

\bibitem[\protect\citeauthoryear{Ling et~al.}{2022}]{ref11}
\begin{botherref}
\oauthor{\bsnm{Ling}, \binits{H.-B.}},
\oauthor{\bsnm{Zhu}, \binits{B.}},
\oauthor{\bsnm{Huang}, \binits{D.}},
\oauthor{\bsnm{Chen}, \binits{D.-H.}},
\oauthor{\bsnm{Wang}, \binits{C.-D.}},
\oauthor{\bsnm{Lai}, \binits{J.-H.}}:
Vision transformer for contrastive clustering.
arXiv:2206.12925
(2022)
\end{botherref}
\endbibitem

\bibitem[\protect\citeauthoryear{Grill et~al.}{2020}]{ref12}
\begin{bchapter}
\bauthor{\bsnm{Grill}, \binits{J.-B.}},
\bauthor{\bsnm{Strub}, \binits{F.}},
\bauthor{\bsnm{Altch\'{e}}, \binits{F.}},
\bauthor{\bsnm{Tallec}, \binits{C.}},
\bauthor{\bsnm{Richemond}, \binits{P.H.}},
\bauthor{\bsnm{Buchatskaya}, \binits{E.}},
\bauthor{\bsnm{Doersch}, \binits{C.}},
\bauthor{\bsnm{Pires}, \binits{B.A.}},
\bauthor{\bsnm{Guo}, \binits{Z.D.}},
\bauthor{\bsnm{Azar}, \binits{M.G.}},
\bauthor{\bsnm{Piot}, \binits{B.}},
\bauthor{\bsnm{Kavukcuoglu}, \binits{K.}},
\bauthor{\bsnm{Munos}, \binits{R.}},
\bauthor{\bsnm{Valko}, \binits{M.}}:
\bctitle{Bootstrap your own latent a new approach to self-supervised learning}.
In: \bbtitle{Proceedings of the 34th International Conference on Neural Information Processing Systems}.
\bsertitle{NIPS '20},
\bconflocation{New York, USA}
(\byear{2020})
\end{bchapter}
\endbibitem

\bibitem[\protect\citeauthoryear{He et~al.}{2020}]{ref13}
\begin{bchapter}
\bauthor{\bsnm{He}, \binits{K.}},
\bauthor{\bsnm{Fan}, \binits{H.}},
\bauthor{\bsnm{Wu}, \binits{Y.}},
\bauthor{\bsnm{Xie}, \binits{S.}},
\bauthor{\bsnm{Girshick}, \binits{R.}}:
\bctitle{Momentum contrast for unsupervised visual representation learning}.
In: \bbtitle{Proceedings of the IEEE/CVF Conference on Computer Vision and Pattern Recognition (CVPR)},
\bconflocation{Seattle, USA}
(\byear{2020}).
\doiurl{10.1109/CVPR42600.2020.00975}
\end{bchapter}
\endbibitem

\bibitem[\protect\citeauthoryear{Wang et~al.}{2021}]{ref32}
\begin{barticle}
\bauthor{\bsnm{Wang}, \binits{B.}},
\bauthor{\bsnm{Ma}, \binits{G.}},
\bauthor{\bsnm{Zhu}, \binits{M.}}:
\batitle{Fast momentum contrast learning for unsupervised person re-identification}.
\bjtitle{IEEE Signal Processing Letters}
\bvolume{28},
\bfpage{2073}--\blpage{2077}
(\byear{2021})
\doiurl{10.1109/LSP.2021.3118564}
\end{barticle}
\endbibitem

\bibitem[\protect\citeauthoryear{Jin et~al.}{2025}]{ref49}
\begin{barticle}
\bauthor{\bsnm{Jin}, \binits{S.}},
\bauthor{\bsnm{Zhou}, \binits{S.}},
\bauthor{\bsnm{Kong}, \binits{D.}},
\bauthor{\bsnm{Han}, \binits{B.}}:
\batitle{Multi-contrast image clustering via multi-resolution augmentation and momentum-output queues}.
\bjtitle{Neurocomputing}
\bvolume{614},
\bfpage{128738}
(\byear{2025})
\doiurl{10.1016/j.neucom.2024.128738}
\end{barticle}
\endbibitem

\bibitem[\protect\citeauthoryear{Ronneberger et~al.}{2015}]{ref14}
\begin{bchapter}
\bauthor{\bsnm{Ronneberger}, \binits{O.}},
\bauthor{\bsnm{Fischer}, \binits{P.}},
\bauthor{\bsnm{Brox}, \binits{T.}}:
\bctitle{U-net: Convolutional networks for biomedical image segmentation}.
In: \bbtitle{Medical Image Computing and Computer-Assisted Intervention -- MICCAI 2015},
pp. \bfpage{234}--\blpage{241}.
\bpublisher{Springer},
\blocation{Cham}
(\byear{2015}).
\doiurl{10.1007/978-3-319-24574-4_28}
\end{bchapter}
\endbibitem

\bibitem[\protect\citeauthoryear{Fu et~al.}{2019}]{ref18}
\begin{bchapter}
\bauthor{\bsnm{Fu}, \binits{J.}},
\bauthor{\bsnm{Liu}, \binits{J.}},
\bauthor{\bsnm{Tian}, \binits{H.}},
\bauthor{\bsnm{Li}, \binits{Y.}},
\bauthor{\bsnm{Bao}, \binits{Y.}},
\bauthor{\bsnm{Fang}, \binits{Z.}},
\bauthor{\bsnm{Lu}, \binits{H.}}:
\bctitle{Dual attention network for scene segmentation}.
In: \bbtitle{Proceedings of the IEEE/CVF Conference on Computer Vision and Pattern Recognition (CVPR)},
\bconflocation{Las Vegas, America}
(\byear{2019}).
\doiurl{10.1109/CVPR.2019.00326}
\end{bchapter}
\endbibitem

\bibitem[\protect\citeauthoryear{Wang et~al.}{2016}]{ref19}
\begin{botherref}
\oauthor{\bsnm{Wang}, \binits{J.}},
\oauthor{\bsnm{Wei}, \binits{Z.}},
\oauthor{\bsnm{Zhang}, \binits{T.}},
\oauthor{\bsnm{Zeng}, \binits{W.}}:
Deeply-fused nets.
arXiv:1605.07716
(2016)
\end{botherref}
\endbibitem

\bibitem[\protect\citeauthoryear{Radford et~al.}{2021}]{ref21}
\begin{bchapter}
\bauthor{\bsnm{Radford}, \binits{A.}},
\bauthor{\bsnm{Kim}, \binits{J.W.}},
\bauthor{\bsnm{Hallacy}, \binits{C.}},
\bauthor{\bsnm{Ramesh}, \binits{A.}},
\bauthor{\bsnm{Goh}, \binits{G.}},
\bauthor{\bsnm{Agarwal}, \binits{S.}},
\bauthor{\bsnm{Sastry}, \binits{G.}},
\bauthor{\bsnm{Askell}, \binits{A.}},
\bauthor{\bsnm{Mishkin}, \binits{P.}},
\bauthor{\bsnm{Clark}, \binits{J.}},
\bauthor{\bsnm{Krueger}, \binits{G.}},
\bauthor{\bsnm{Sutskever}, \binits{I.}}:
\bctitle{Learning transferable visual models from natural language supervision}.
In: \bbtitle{Proceedings of the 38th International Conference on Machine Learning}.
\bsertitle{Proceedings of Machine Learning Research},
vol. \bseriesno{139},
pp. \bfpage{8748}--\blpage{8763}
(\byear{2021})
\end{bchapter}
\endbibitem

\bibitem[\protect\citeauthoryear{Li et~al.}{2022}]{ref48}
\begin{bchapter}
\bauthor{\bsnm{Li}, \binits{L.H.}},
\bauthor{\bsnm{Zhang}, \binits{P.}},
\bauthor{\bsnm{Zhang}, \binits{H.}},
\bauthor{\bsnm{Yang}, \binits{J.}},
\bauthor{\bsnm{Li}, \binits{C.}},
\bauthor{\bsnm{Zhong}, \binits{Y.}},
\bauthor{\bsnm{Wang}, \binits{L.}},
\bauthor{\bsnm{Yuan}, \binits{L.}},
\bauthor{\bsnm{Zhang}, \binits{L.}},
\bauthor{\bsnm{Hwang}, \binits{J.-N.}},
\bauthor{\bsnm{Chang}, \binits{K.-W.}},
\bauthor{\bsnm{Gao}, \binits{J.}}:
\bctitle{Grounded language-image pre-training}.
In: \bbtitle{2022 IEEE/CVF Conference on Computer Vision and Pattern Recognition (CVPR)},
pp. \bfpage{10955}--\blpage{10965}
(\byear{2022}).
\doiurl{10.1109/CVPR52688.2022.01069}
\end{bchapter}
\endbibitem

\bibitem[\protect\citeauthoryear{Cao et~al.}{2022}]{ref43}
\begin{barticle}
\bauthor{\bsnm{Cao}, \binits{W.}},
\bauthor{\bsnm{Wu}, \binits{Y.}},
\bauthor{\bsnm{Huang}, \binits{C.}},
\bauthor{\bsnm{Patwary}, \binits{M.J.A.}},
\bauthor{\bsnm{Wang}, \binits{X.}}:
\batitle{Mff: Multi-modal feature fusion for zero-shot learning}.
\bjtitle{Neurocomputing}
\bvolume{510},
\bfpage{172}--\blpage{180}
(\byear{2022})
\doiurl{10.1016/j.neucom.2022.09.070}
\end{barticle}
\endbibitem

\bibitem[\protect\citeauthoryear{Dosovitskiy et~al.}{2021}]{ref20}
\begin{bchapter}
\bauthor{\bsnm{Dosovitskiy}, \binits{A.}},
\bauthor{\bsnm{Beyer}, \binits{L.}},
\bauthor{\bsnm{Kolesnikov}, \binits{A.}},
\bauthor{\bsnm{Weissenborn}, \binits{D.}},
\bauthor{\bsnm{Zhai}, \binits{X.}},
\bauthor{\bsnm{Unterthiner}, \binits{T.}},
\bauthor{\bsnm{Dehghani}, \binits{M.}},
\bauthor{\bsnm{Minderer}, \binits{M.}},
\bauthor{\bsnm{Heigold}, \binits{G.}},
\bauthor{\bsnm{Gelly}, \binits{S.}}, \betal:
\bctitle{An image is worth 16x16 words: Transformers for image recognition at scale}.
In: \bbtitle{In International Conference on Learning Representations(ICLR)}
(\byear{2021})
\end{bchapter}
\endbibitem

\bibitem[\protect\citeauthoryear{Hu et~al.}{2023}]{ref39}
\begin{barticle}
\bauthor{\bsnm{Hu}, \binits{Z.}},
\bauthor{\bsnm{Wang}, \binits{Y.}},
\bauthor{\bsnm{Ning}, \binits{H.}},
\bauthor{\bsnm{Wu}, \binits{D.}},
\bauthor{\bsnm{Nie}, \binits{F.}}:
\batitle{Mutual-taught deep clustering}.
\bjtitle{Knowledge-Based Systems}
\bvolume{282},
\bfpage{111100}
(\byear{2023})
\doiurl{10.1016/j.knosys.2023.111100}
\end{barticle}
\endbibitem

\bibitem[\protect\citeauthoryear{Wu et~al.}{2023}]{ref58}
\begin{barticle}
\bauthor{\bsnm{Wu}, \binits{L.}},
\bauthor{\bsnm{Zhang}, \binits{W.}},
\bauthor{\bsnm{Jiang}, \binits{T.}},
\bauthor{\bsnm{Yang}, \binits{W.}},
\bauthor{\bsnm{Jin}, \binits{X.}},
\bauthor{\bsnm{Zeng}, \binits{W.}}:
\batitle{[CLS] Token is All You Need for Zero-Shot Semantic Segmentation}.
\bjtitle{arXiv preprint arXiv:2304.06212},
\byear{2023}
\end{barticle}
\endbibitem

\bibitem[\protect\citeauthoryear{Yoo et~al.}{2023}]{ref59}
\begin{barticle}
\bauthor{\bsnm{Yoo}, \binits{S.}},
\bauthor{\bsnm{Kim}, \binits{E.}},
\bauthor{\bsnm{Jung}, \binits{D.}},
\bauthor{\bsnm{Lee}, \binits{J.}},
\bauthor{\bsnm{Yoon}, \binits{S.}}:
\batitle{Improving Visual Prompt Tuning for Self-supervised Vision Transformers}.
\bjtitle{Proceedings of the 40th International Conference on Machine Learning},
\bvolume{202},
\bfpage{40075}--\blpage{40092},
\byear{2023},
\url{https://proceedings.mlr.press/v202/yoo23a.html}
\end{barticle}
\endbibitem

\bibitem[\protect\citeauthoryear{Zhou et~al.}{2022}]{ref60}
\begin{barticle}
\bauthor{\bsnm{Zhou}, \binits{K.}},
\bauthor{\bsnm{Yang}, \binits{J.}},
\bauthor{\bsnm{Loy}, \binits{C.~C.}},
\bauthor{\bsnm{Liu}, \binits{Z.}}:
\batitle{Conditional Prompt Learning for Vision-Language Models}.
\bjtitle{Proceedings of the IEEE/CVF Conference on Computer Vision and Pattern Recognition (CVPR)},
\bfpage{16816}--\blpage{16825},
\byear{2022},
\doiurl{10.1109/CVPR52688.2022.01631}
\end{barticle}
\endbibitem

\bibitem[\protect\citeauthoryear{Lecun et~al.}{1998}]{ref23}
\begin{barticle}
\bauthor{\bsnm{Lecun}, \binits{Y.}},
\bauthor{\bsnm{Bottou}, \binits{L.}},
\bauthor{\bsnm{Bengio}, \binits{Y.}},
\bauthor{\bsnm{Haffner}, \binits{P.}}:
\batitle{Gradient-based learning applied to document recognition}.
\bjtitle{Proceedings of the IEEE}
\bvolume{86}(\bissue{11}),
\bfpage{2278}--\blpage{2324}
(\byear{1998})
\doiurl{10.1109/5.726791}
\end{barticle}
\endbibitem

\bibitem[\protect\citeauthoryear{He et~al.}{2016}]{ref17}
\begin{bchapter}
\bauthor{\bsnm{He}, \binits{K.}},
\bauthor{\bsnm{Zhang}, \binits{X.}},
\bauthor{\bsnm{Ren}, \binits{S.}},
\bauthor{\bsnm{Sun}, \binits{J.}}:
\bctitle{Deep residual learning for image recognition}.
In: \bbtitle{Proceedings of the IEEE Conference on Computer Vision and Pattern Recognition (CVPR)},
\bconflocation{Las Vegas, America}
(\byear{2016}).
\doiurl{10.1109/CVPR.2016.90}
\end{bchapter}
\endbibitem

\bibitem[\protect\citeauthoryear{Li et~al.}{2022}]{ref54}
\begin{bchapter}
\bauthor{\bsnm{Li}, \binits{Z.}},
\bauthor{\bsnm{Xu}, \binits{B.}},
\bauthor{\bsnm{Zhu}, \binits{C.}},
\bauthor{\bsnm{Zhao}, \binits{T.}}:
\bctitle{{CLMLF}: A Contrastive Learning and Multi-Layer Fusion Method for Multimodal Sentiment Detection}.
In: \bbtitle{Findings of the Association for Computational Linguistics: NAACL 2022},
\beditor{\bparticle{Editors: }\bauthor{\bsnm{Carpuat}, \binits{M.}},
\bauthor{\bsnm{de Marneffe}, \binits{M.-C.}},
\bauthor{\bsnm{Meza Ruiz}, \binits{I. V.}}},
\bconflocation{Seattle, United States}
(\byear{2022}),
\bfpage{2282}--\blpage{2294}.
\doiurl{10.18653/v1/2022.findings-naacl.175}.
\end{bchapter}
\endbibitem


\bibitem[\protect\citeauthoryear{Huang et~al.}{2017}]{ref55}
\begin{barticle}
\bauthor{\bsnm{Huang}, \binits{G.}},
\bauthor{\bsnm{Liu}, \binits{Z.}},
\bauthor{\bsnm{van der Maaten}, \binits{L.}},
\bauthor{\bsnm{Weinberger}, \binits{K. Q.}}:
\batitle{Densely Connected Convolutional Networks}.
\bjtitle{Proceedings of the IEEE Conference on Computer Vision and Pattern Recognition (CVPR)},
\bfpage{2261}--\blpage{2269}
(\byear{2017})
\doiurl{10.1109/CVPR.2017.243}
\end{barticle}
\endbibitem

\bibitem[\protect\citeauthoryear{Li \& Xu}{2023}]{ref56}
\begin{barticle}
\bauthor{\bsnm{Li}, \binits{Y.} \binits{X.}},
\bauthor{\bsnm{Xu}, \binits{C.}}:
\batitle{Trade-Off Between Robustness and Accuracy of Vision Transformers}.
\bjtitle{Proceedings of the IEEE/CVF Conference on Computer Vision and Pattern Recognition (CVPR)},
\bfpage{7558}--\blpage{7568}
(\byear{2023})
\end{barticle}
\endbibitem

\bibitem[\protect\citeauthoryear{Wang et~al.}{2022}]{ref57}
\begin{barticle}
\bauthor{\bsnm{Wang}, \binits{Y.}},
\bauthor{\bsnm{Chen}, \binits{X.}},
\bauthor{\bsnm{Cao}, \binits{L.}},
\bauthor{\bsnm{Huang}, \binits{W.}},
\bauthor{\bsnm{Sun}, \binits{F.}},
\bauthor{\bsnm{Wang}, \binits{Y.}}:
\batitle{Multimodal Token Fusion for Vision Transformers}.
\bjtitle{Proceedings of the IEEE/CVF Conference on Computer Vision and Pattern Recognition (CVPR)},
\bfpage{12186}--\blpage{12195}
(\byear{2022})
\end{barticle}
\endbibitem

\bibitem[\protect\citeauthoryear{Oord et~al.}{2018}]{ref35}
\begin{botherref}
\oauthor{\bsnm{Oord}, \binits{A.v.d.}},
\oauthor{\bsnm{Li}, \binits{Y.}},
\oauthor{\bsnm{Vinyals}, \binits{O.}}:
Representation learning with contrastive predictive coding.
arXiv:1807.03748
(2018)
\end{botherref}
\endbibitem

\bibitem[\protect\citeauthoryear{Devlin}{2018}]{ref45}
\begin{botherref}
\oauthor{\bsnm{Devlin}, \binits{J.}}:
Bert: Pre-training of deep bidirectional transformers for language understanding.
arXiv:1810.04805
(2018)
\end{botherref}
\endbibitem

\bibitem[\protect\citeauthoryear{Long et~al.}{2017}]{ref24}
\begin{barticle}
\bauthor{\bsnm{Long}, \binits{Y.}},
\bauthor{\bsnm{Gong}, \binits{Y.}},
\bauthor{\bsnm{Xiao}, \binits{Z.}},
\bauthor{\bsnm{Liu}, \binits{Q.}}:
\batitle{Accurate object localization in remote sensing images based on convolutional neural networks}.
\bjtitle{IEEE Transactions on Geoscience and Remote Sensing}
\bvolume{55}(\bissue{5}),
\bfpage{2486}--\blpage{2498}
(\byear{2017})
\doiurl{10.1109/TGRS.2016.2645610}
\end{barticle}
\endbibitem

\bibitem[\protect\citeauthoryear{Yang and Newsam}{2010}]{ref25}
\begin{bchapter}
\bauthor{\bsnm{Yang}, \binits{Y.}},
\bauthor{\bsnm{Newsam}, \binits{S.}}:
\bctitle{Bag-of-visual-words and spatial extensions for land-use classification}.
In: \bbtitle{Proceedings of the 18th SIGSPATIAL International Conference on Advances in Geographic Information Systems},
pp. \bfpage{270}--\blpage{279}.
\bpublisher{Association for Computing Machinery},
\blocation{New York, NY, USA}
(\byear{2010}).
\doiurl{10.1145/1869790.1869829}
\end{bchapter}
\endbibitem

\bibitem[\protect\citeauthoryear{Zhao et~al.}{2016}]{ref26}
\begin{barticle}
\bauthor{\bsnm{Zhao}, \binits{B.}},
\bauthor{\bsnm{Zhong}, \binits{Y.}},
\bauthor{\bsnm{Xia}, \binits{G.-S.}},
\bauthor{\bsnm{Zhang}, \binits{L.}}:
\batitle{Dirichlet-derived multiple topic scene classification model for high spatial resolution remote sensing imagery}.
\bjtitle{IEEE Transactions on Geoscience and Remote Sensing}
\bvolume{54}(\bissue{4}),
\bfpage{2108}--\blpage{2123}
(\byear{2016})
\doiurl{10.1109/TGRS.2015.2496185}
\end{barticle}
\endbibitem

\bibitem[\protect\citeauthoryear{Xia et~al.}{2017}]{ref27}
\begin{barticle}
\bauthor{\bsnm{Xia}, \binits{G.-S.}},
\bauthor{\bsnm{Hu}, \binits{J.}},
\bauthor{\bsnm{Hu}, \binits{F.}},
\bauthor{\bsnm{Shi}, \binits{B.}},
\bauthor{\bsnm{Bai}, \binits{X.}},
\bauthor{\bsnm{Zhong}, \binits{Y.}},
\bauthor{\bsnm{Zhang}, \binits{L.}},
\bauthor{\bsnm{Lu}, \binits{X.}}:
\batitle{Aid: A benchmark data set for performance evaluation of aerial scene classification}.
\bjtitle{IEEE Transactions on Geoscience and Remote Sensing}
\bvolume{55}(\bissue{7}),
\bfpage{3965}--\blpage{3981}
(\byear{2017})
\doiurl{10.1109/TGRS.2017.2685945}
\end{barticle}
\endbibitem

\bibitem[\protect\citeauthoryear{Chang et~al.}{2017}]{ref41}
\begin{bchapter}
\bauthor{\bsnm{Chang}, \binits{J.}},
\bauthor{\bsnm{Wang}, \binits{L.}},
\bauthor{\bsnm{Meng}, \binits{G.}},
\bauthor{\bsnm{Xiang}, \binits{S.}},
\bauthor{\bsnm{Pan}, \binits{C.}}:
\bctitle{Deep adaptive image clustering}.
In: \bbtitle{2017 IEEE International Conference on Computer Vision (ICCV)},
pp. \bfpage{5880}--\blpage{5888}
(\byear{2017}).
\doiurl{10.1109/ICCV.2017.626}
\end{bchapter}
\endbibitem

\bibitem[\protect\citeauthoryear{Krizhevsky~A.}{2009}]{ref40}
\begin{bbook}
\bauthor{\bsnm{Krizhevsky~A.}, \binits{e.a.} \bsuffix{Hinton~G.}}:
\bbtitle{Learning Multiple Layers of Features from Tiny Images},
(\byear{2009})
\end{bbook}
\endbibitem

\bibitem[\protect\citeauthoryear{Strehl and Ghosh}{2002}]{ref36}
\begin{barticle}
\bauthor{\bsnm{Strehl}, \binits{A.}},
\bauthor{\bsnm{Ghosh}, \binits{J.}}:
\batitle{Cluster ensembles---a knowledge reuse framework for combining multiple partitions}.
\bjtitle{Journal of machine learning research}
\bvolume{3}(\bissue{Dec}),
\bfpage{583}--\blpage{617}
(\byear{2002})
\doiurl{10.1162/153244303321897735}
\end{barticle}
\endbibitem

\bibitem[\protect\citeauthoryear{Huang et~al.}{2020}]{ref37}
\begin{barticle}
\bauthor{\bsnm{Huang}, \binits{D.}},
\bauthor{\bsnm{Wang}, \binits{C.-D.}},
\bauthor{\bsnm{Wu}, \binits{J.-S.}},
\bauthor{\bsnm{Lai}, \binits{J.-H.}},
\bauthor{\bsnm{Kwoh}, \binits{C.-K.}}:
\batitle{Ultra-scalable spectral clustering and ensemble clustering}.
\bjtitle{IEEE Transactions on Knowledge and Data Engineering}
\bvolume{32}(\bissue{6}),
\bfpage{1212}--\blpage{1226}
(\byear{2020})
\doiurl{10.1109/TKDE.2019.2903410}
\end{barticle}
\endbibitem

\bibitem[\protect\citeauthoryear{Huang et~al.}{2021}]{ref38}
\begin{barticle}
\bauthor{\bsnm{Huang}, \binits{D.}},
\bauthor{\bsnm{Wang}, \binits{C.-D.}},
\bauthor{\bsnm{Lai}, \binits{J.-H.}},
\bauthor{\bsnm{Kwoh}, \binits{C.-K.}}:
\batitle{Toward multidiversified ensemble clustering of high-dimensional data: From subspaces to metrics and beyond}.
\bjtitle{IEEE Transactions on Cybernetics}
\bvolume{52}(\bissue{11}),
\bfpage{12231}--\blpage{12244}
(\byear{2021})
\doiurl{10.1109/TCYB.2021.3049633}
\end{barticle}
\endbibitem

\bibitem[\protect\citeauthoryear{Touvron et~al.}{2021}]{ref28}
\begin{botherref}
\oauthor{\bsnm{Touvron}, \binits{H.}},
\oauthor{\bsnm{Cord}, \binits{M.}},
\oauthor{\bsnm{Douze}, \binits{M.}},
\oauthor{\bsnm{Massa}, \binits{F.}},
\oauthor{\bsnm{Sablayrolles}, \binits{A.}},
\oauthor{\bsnm{Jegou}, \binits{H.}}:
Training data-efficient image transformers \& amp; distillation through attention
\textbf{139},
10347--10357
(2021)
\end{botherref}
\endbibitem

\bibitem[\protect\citeauthoryear{Xie et~al.}{2016}]{ref30}
\begin{barticle}
\bauthor{\bsnm{Xie}, \binits{J.}},
\bauthor{\bsnm{Girshick}, \binits{R.}},
\bauthor{\bsnm{Farhadi}, \binits{A.}}:
\batitle{Unsupervised deep embedding for clustering analysis}.
\bjtitle{Proceedings of The 33rd International Conference on Machine Learning, PMLR}
\bvolume{48},
\bfpage{478}--\blpage{487}
(\byear{2016})
\end{barticle}
\endbibitem

\bibitem[\protect\citeauthoryear{Tao et~al.}{2021}]{ref31}
\begin{botherref}
\oauthor{\bsnm{Tao}, \binits{Y.}},
\oauthor{\bsnm{Takagi}, \binits{K.}},
\oauthor{\bsnm{Nakata}, \binits{K.}}:
Clustering-friendly representation learning via instance discrimination and feature decorrelation.
arXiv:2106.00131
(2021)
\end{botherref}
\endbibitem

\bibitem[\protect\citeauthoryear{Wei et~al.}{2024}]{ref61}
\begin{barticle}
\bauthor{\bsnm{Wei}, \binits{X.}},
\bauthor{\bsnm{Hu}, \binits{T.}},
\bauthor{\bsnm{Meng}, \binits{L.}},
\bauthor{\bsnm{Zhao}, \binits{C.}},
\bauthor{\bsnm{Yang}, \binits{F.}},
\bauthor{\bsnm{Wei}, \binits{Z.}},
\bauthor{\bsnm{Lu}, \binits{Y.}}:
\batitle{PDTE: Pseudo Dual-View Contrastive Learning via Taylor Expansion Optimization for Remote Sensing Image Clustering}.
\bjtitle{IEEE Sensors Journal}
\bvolume{24}(\bissue{20}),
\bfpage{32813}--\blpage{32821}
(\byear{2024})
\doiurl{10.1109/JSEN.2024.3452070}
\end{barticle}
\endbibitem

\bibitem[\protect\citeauthoryear{Cai et~al.}{2023}]{ref62}
\begin{barticle}
\bauthor{\bsnm{Cai}, \binits{S.}},
\bauthor{\bsnm{Qiu}, \binits{L.}},
\bauthor{\bsnm{Chen}, \binits{X.}},
\bauthor{\bsnm{Zhang}, \binits{Q.}},
\bauthor{\bsnm{Chen}, \binits{L.}}:
\batitle{Semantic-enhanced image clustering}.
\bjtitle{Proceedings of the Thirty-Seventh AAAI Conference on Artificial Intelligence and Thirty-Fifth Conference on Innovative Applications of Artificial Intelligence and Thirteenth Symposium on Educational Advances in Artificial Intelligence},
\bvolume{37},
\bfpage{6869}--\blpage{6878}
(\byear{2023}).
\doiurl{10.1609/aaai.v37i6.25841}
\end{barticle}
\endbibitem

\end{thebibliography}

\end{document}